%% file: main.tex
\documentclass[10pt,twocolumn,letterpaper]{article}

\usepackage{cvpr}              % To produce the CAMERA-READY version
\input{preamble}

\definecolor{cvprblue}{rgb}{0.21,0.49,0.74}
\usepackage[pagebackref,breaklinks,colorlinks,allcolors=cvprblue]{hyperref}

\title{UnicEdit-10M: A Dataset and Benchmark Breaking the Scale-Quality Barrier via Unified Verification for Reasoning-Enriched Edits}

\author{
  Keming Ye$^{1}$\footnotemark[1] \quad
  Zhipeng Huang$^{2}$ \quad
  Canmiao Fu$^{2}$ \quad
  Qingyang Liu$^{3}$\footnotemark[1] \quad
  Jiani Cai$^{4}$ \quad \\
  Zheqi Lv$^{1}$ \quad
  Chen Li$^{2}$ \quad
  Jing LYU$^{2}$ \quad
  Zhou Zhao$^{1}$ \quad
  Shengyu Zhang$^{1}$\footnotemark[2] \quad \\[0.2cm]
  $^{1}$Zhejiang University \quad
  $^{2}$WeChat Vision, Tencent Inc. \quad \\
  $^{3}$Shanghai Jiao Tong University \quad
  $^{4}$Xinjiang University \\[0.5cm] % 这里添加垂直间距，与单位隔开
  \href{https://hongsexiaotanhua.github.io/UnicEdit-10M/}{\color{black}\faGlobe\ Project Page} \qquad
  \href{https://github.com/WeChatCV/UnicBench}{\color{black}\faGithub\ GitHub Code} \qquad
\href{https://huggingface.co/datasets/xiaotanhua/UnicBench}{\color{black}\hflogo\ Benchmark}
}

\usepackage[most]{tcolorbox} 
\tcbuselibrary{listings}

\usepackage[capitalize]{cleveref}

\crefname{promptlisting}{Prompt}{Prompts}
\Crefname{promptlisting}{Prompt}{Prompts}

\newcounter{promptlisting}

\newtcblisting[
  use counter=promptlisting,
]{promptlisting}[1][]{%
  enhanced,%
  breakable,%
  width=\textwidth,%
  colback=gray!5,%
  colframe=black!70,%
  fonttitle=\bfseries,%
  title={Prompt~\thetcbcounter},%
  listing only,%
  listing options={%
    basicstyle=\ttfamily\small,
    breaklines=true,
    breakindent=0pt,
    columns=fullflexible
  },%
  #1%
}

\begin{document}
% \maketitle

% \begin{figure*}[ht]
% \centering
% \includegraphics[width=\linewidth]{ImageEditBench/figures/teaser_v2p3.pdf}
% \end{figure*}

% \begin{strip}
%   \centering
%   \includegraphics[width=\textwidth]{figures/teaser_v2p3.pdf}
%   \captionof{figure}{teaser}
% \end{strip}

\twocolumn[{%
	\renewcommand\twocolumn[1][]{#1}%
	\maketitle
	\begin{center}
		\centering
		\captionsetup{type=figure}
		\includegraphics[width=1\linewidth]{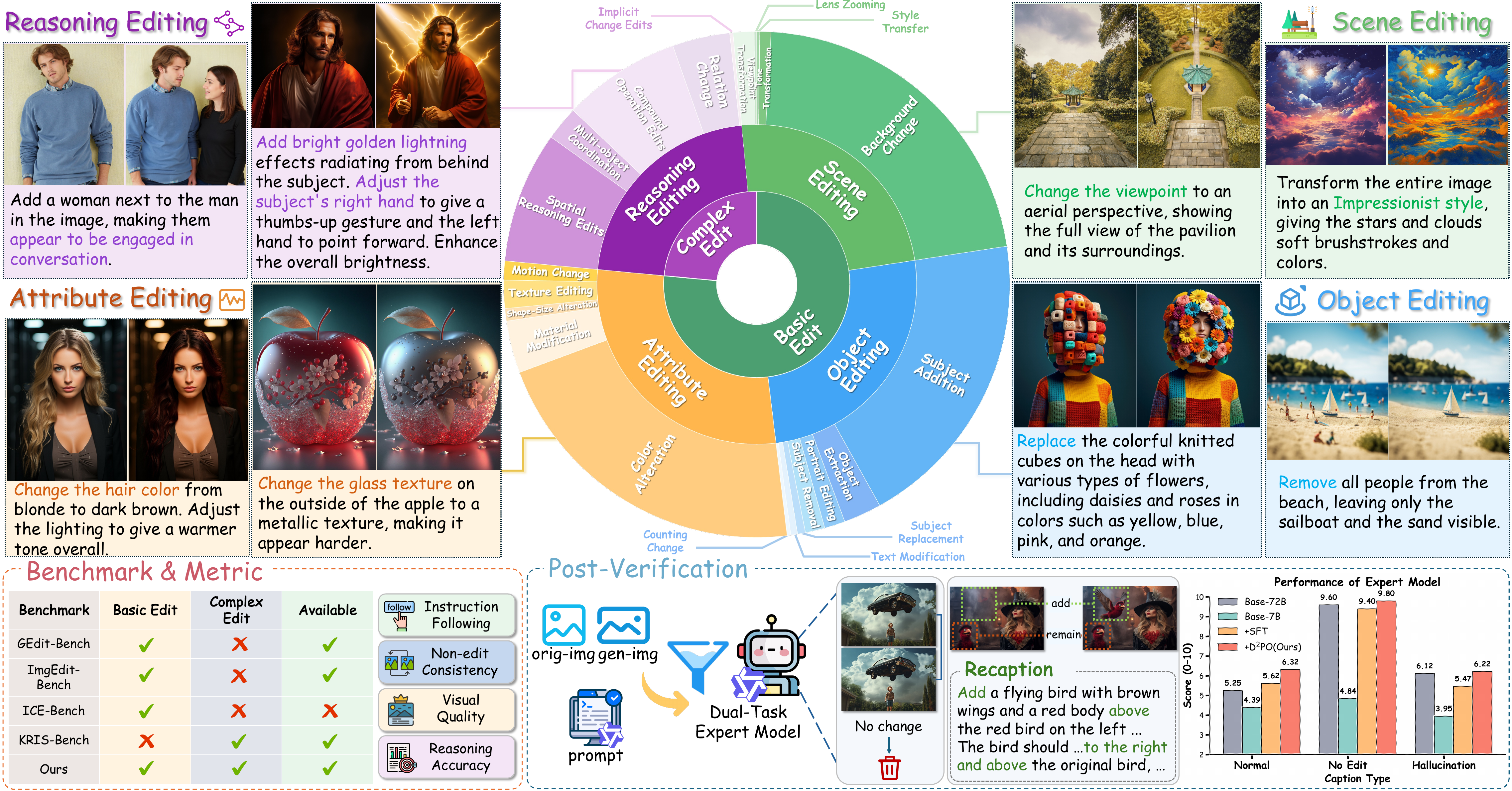}
        % \vspace{-0.1cm}
		\captionof{figure}{\textbf{UnicEdit-10M} covers 22 edit tasks spanning basic and complex edits, with a unified post-verification stage that filters failures and refines instructions to yield high-quality triplets. We also introduce \textbf{UnicBench} with fine-grained metrics for comprehensive evaluation.}
        % \vspace{-0.1cm}
		\label{fig:teaser}
	\end{center}
}]

\begin{figure*}[t]
    \centering
    \includegraphics[width=1\textwidth]{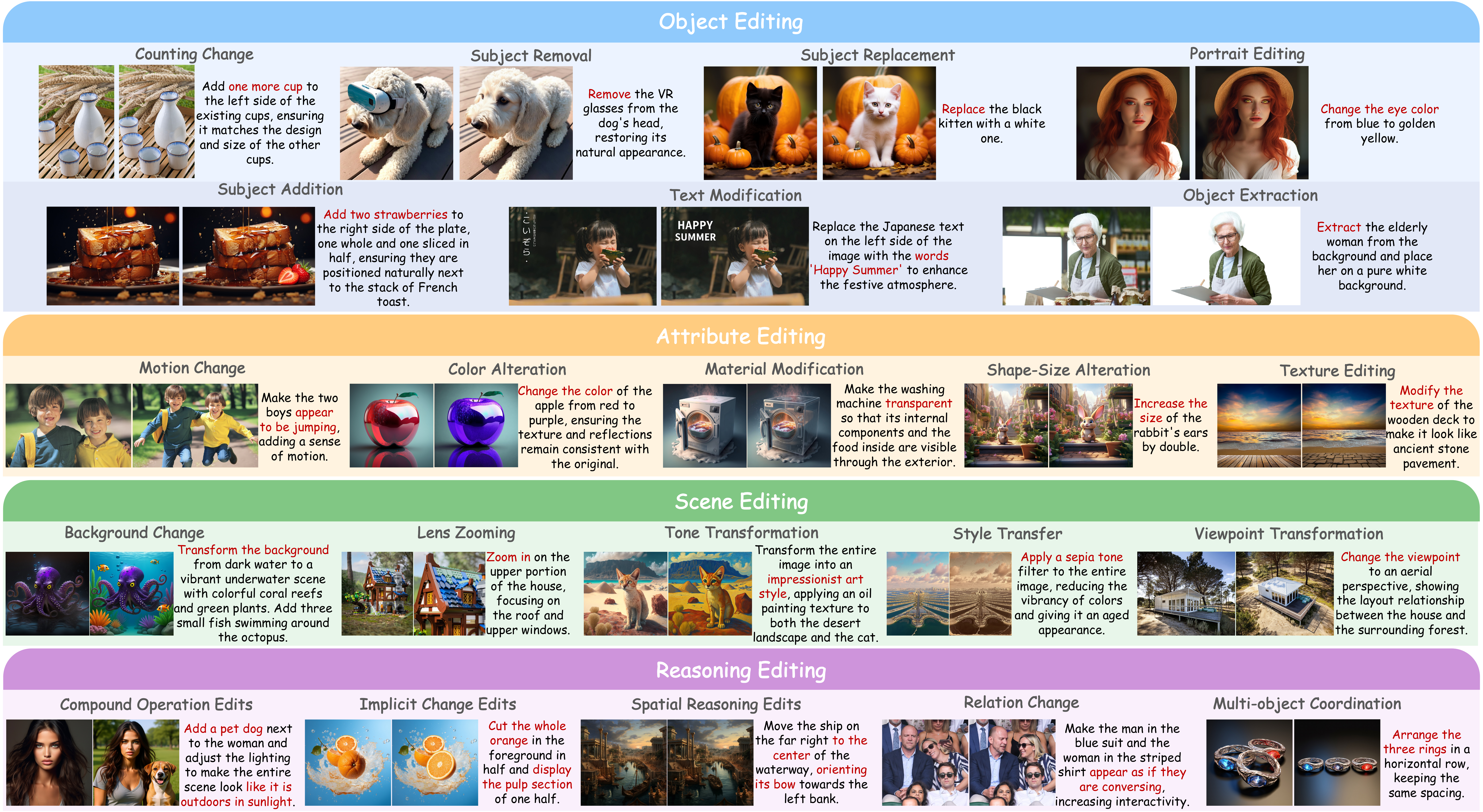}
    \vspace{-0.5cm}
    \caption{Representative examples of all sub-tasks from \textbf{UnicEdit-10M}.}
    \vspace{-0.5cm}
    \label{fig:showcases}
\end{figure*}

\renewcommand{\thefootnote}{\fnsymbol{footnote}}
\footnotetext[1]{Work done during an internship at WeChat Vision, Tencent Inc.}
\footnotetext[2]{Corresponding author.}

\input{tex/0abstract}
\input{tex/1introduction}
\input{tex/2related_works}
\input{tex/3methodology}
\input{tex/4experiments}
\input{tex/5conclusion}

% \newpage
{
    \small
    \bibliographystyle{ieeenat_fullname}
    \bibliography{main}
}

\input{tex/9suppl}

\end{document}

%% file: preamble.tex
\newcommand{\red}[1]{{\color{red}#1}}

\usepackage{amsmath}
\usepackage{amsthm}
\usepackage{graphicx}
\usepackage[table,xcdraw]{xcolor}

\usepackage[utf8]{inputenc}
\usepackage{booktabs}   % \toprule, \midrule, \bottomrule, \specialrule
\usepackage{pifont}     % \ding for check / cross
\usepackage{array}      % column formatting

\usepackage{pifont}   % For checkmarks and crosses
\usepackage{multirow} % For multi-row cells in tables
\usepackage{mathtools}
\usepackage{algorithm}% http://ctan.org/pkg/algorithms
\usepackage{algpseudocode}% http://ctan.org/pkg/algorithmicx

\newcommand{\cmark}{\textcolor{green!60!black}{\ding{52}}}
\newcommand{\xmark}{\textcolor{red}{\ding{56}}}

\newcommand{\DPO}{$\text{D}^{2}\text{PO}$}

\usepackage{caption}
\usepackage{cuted}
\usepackage{capt-of} % 提供 captionof，用于在 \twocolumn[] 中放 caption

\usepackage[normalem]{ulem}
\useunder{\uline}{\ul}{}

\usepackage{makecell}

\usepackage{fontawesome5}  % 用于图标 (如地球仪、Github猫、奖杯等)
\usepackage{calc}     % 用于计算高度，可选但推荐

\newcommand{\hflogo}{\raisebox{-0.75ex}{\includegraphics[height=1.4em]{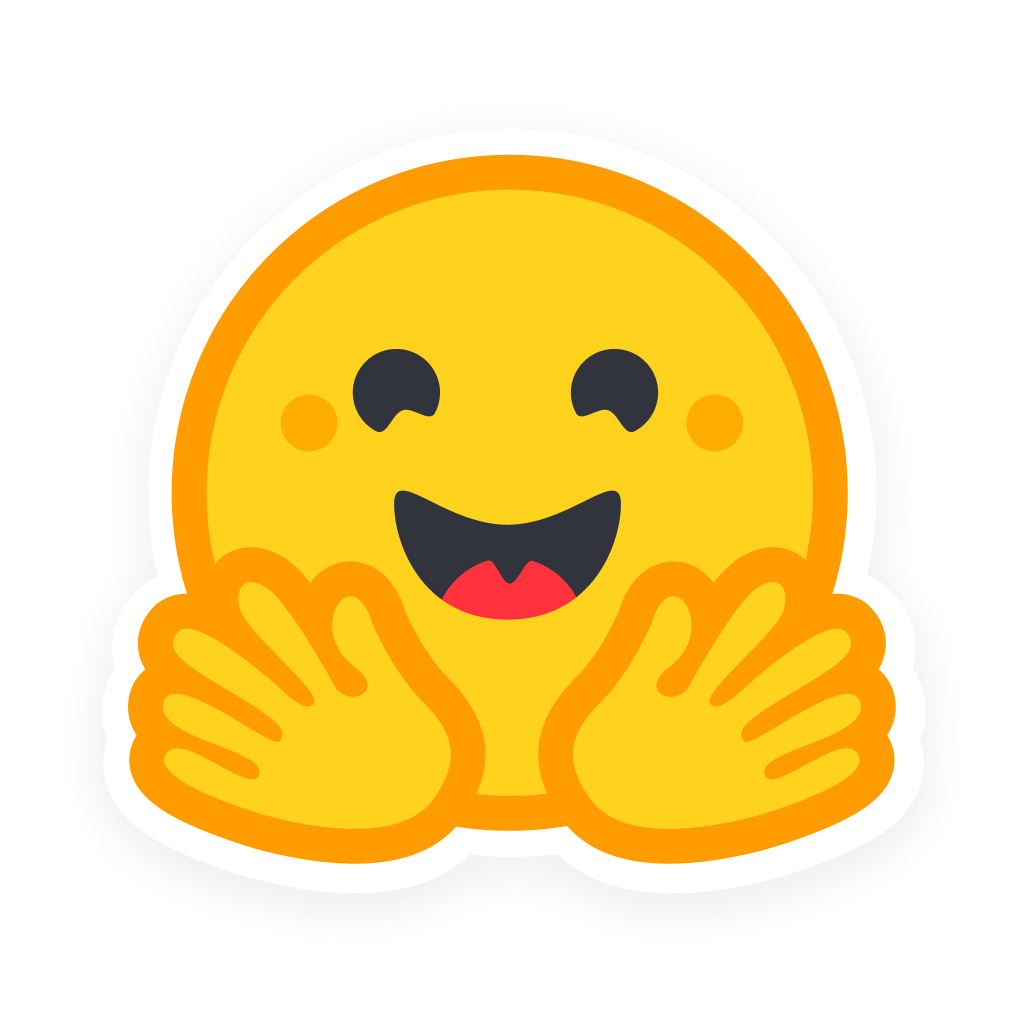}}}

%% file: tex/0abstract.tex
\begin{abstract}
With the rapid advances of powerful multimodal models such as GPT-4o, Nano Banana, and Seedream 4.0 in Image Editing, the performance gap between closed-source and open-source models is widening, primarily due to the scarcity of large-scale, high-quality training data and comprehensive benchmarks capable of diagnosing model weaknesses across diverse editing behaviors. Existing data construction methods face a scale-quality trade-off: human annotations are high-quality but not scalable, while automated pipelines suffer from error propagation and noise. To address this, we introduce a lightweight data pipeline that replaces multi-toolchains with an end-to-end model and a unified post-verification stage. For scalable quality control, we train a 7B dual-task expert model, \textbf{Qwen-Verify}, for efficient failure detection and instruction recaptioning. This pipeline yields \textbf{UnicEdit-10M}, a 10M-scale dataset spanning diverse basic and complex editing tasks. We also propose \textbf{UnicBench}, a general benchmark that extends beyond basic edits to explicitly assess spatial and knowledge-driven reasoning. To enable fine-grained diagnosis, we introduce novel metrics, including \textit{Non-edit Consistency} and \textit{Reasoning Accuracy}. Our analysis of mainstream models on UnicBench reveals their limitations and provides clear directions for future research.
\end{abstract}

%% file: tex/1introduction.tex
\section{Introduction}
Instruction-based image editing~\cite{huang2025survey-edit,shuai2024survey-edit} has progressed rapidly, enabling complex visual edits from natural-language instructions. Closed-source models such as GPT-4o~\cite{hurst2024gpt4o}, Nano Banana~\cite{comanici2025nanobanana}, and Seedream 4.0~\cite{seedream2025seedream4p0} exemplify this progress, showcasing capabilities in understanding nuanced commands and producing semantically consistent edits. 
Yet the gap between closed- and open-source models continues to widen, primarily due to the lack of two critical components: (1) large-scale, high-quality public datasets for training, and (2) comprehensive benchmarks capable of diagnosing model weaknesses across diverse editing tasks.

However, existing open-source datasets and benchmarks face significant limitations. Current datasets present a scale-quality trade-off: human-annotated collections~\cite{zhang2023magicbrush} offer high fidelity but are small, while large-scale automated pipelines introduce systematic noise like mismatched instructions and failed edits~\cite{ye2025imgedit,ge2024seeddataedit}. On the benchmark side, prevailing evaluations emphasize basic edits~\cite{basu2023editval,pan2025icebench,ma2024i2ebench} and offer sparse coverage of complex reasoning tasks, limiting their ability to diagnose higher-order model weaknesses.

We attribute these limitations to three technical causes:
(1) \textbf{Error propagation from complex toolchains.} Automated pipelines that rely on complex toolchains~\cite{liu2025step1x, ye2025imgedit} are prone to error propagation, where minor inaccuracies in early stages amplify into significant artifacts downstream.
(2) \textbf{Insufficient or narrowly scoped post-verification.} While some automated methods incorporate post-verification, these steps are often one-dimensional and prohibitively expensive. For instance, some approaches~\cite{kuprashevich2025nohumansrequired} perform simple failure detection without correcting semantic misalignments, while others~\cite{wang2025gptimageedit} focus solely on instruction refinement using the GPT-4o~\cite{hurst2024gpt4o} API at a high cost, overlooking image quality.
(3) \textbf{Evaluation blind spots for complex edits.} Existing benchmarks often emphasize object- and attribute-level changes, lacking systematic tests for complex reasoning and spatial understanding~\cite{liu2025step1x,ye2025imgedit}. Moreover, current VLM-based metrics may ignore unintended changes in non-edited regions and tend to be sensitive to stylistic variations when judging naturalness, preventing a comprehensive assessment of higher-order editing skills.

\input{tables/tab_datasets}
To overcome these limitations, we introduce a scalable pipeline that curates a 10M \underline{Uni}versal Dataset for \underline{c}omprehensive image editing (UnicEdit-10M), spanning basic attribute/object edits, spatial viewpoint transformations, and reasoning-driven tasks. We further propose a general benchmark that systematically evaluates both basic and complex editing capabilities.
Our pipeline consists of three stages: First, we leverage open-source VLMs to generate diverse editing instructions. Second, to circumvent error propagation, we employ an end-to-end editing model for instruction-based image editing, instead of multi-tool invocation. Third, all synthesized ⟨\textit{original, instruction, edited}⟩ triplets pass through a unified post-verification stage, which not only filters out faulty cases but also refines the corresponding instructions to enhance semantic alignment with the visual edits. To make verification practical at scale, we train a compact 7B dual-task expert model Qwen-Verify that jointly (1) performs fine-grained failure detection and (2) generates instruction recaptions, providing efficient, post-hoc validation and low-cost correction without relying on expensive, large API calls.
To fix evaluation blind spots, we construct UnicBench, a benchmark that extends beyond conventional basic edits to include spatial viewpoint changes, multi-object coordination, and knowledge-driven reasoning, collectively referred to as \textit{complex edits}. 
For precise evaluation, we augment the VIEScore~\cite{ku2023viescore} with a \textit{Non-edit Consistency} metric that detects unexpected changes in unmodified regions, and propose \textit{Reasoning Accuracy}, which evaluates whether the semantic or causal outcomes implied by an instruction are faithfully realized in the edited image.

In summary, our contributions are threefold:
% \begin{enumerate}
\begin{itemize}
    \item We propose a scalable, quality-aware data curation pipeline that yields \textbf{UnicEdit-10M}, a 10M dataset, extending beyond basic edits to cover complex spatial, viewpoint transformation and reasoning edits, achieving the SOTA aesthetic quality across other datasets.
    \item We introduce \textbf{Qwen-Verify}, a 7B dual-task expert model that performs failure detection and instruction recaptioning. It achieves this with high efficiency, outperforming Qwen2.5-VL-72B at a fraction of the computational and economic cost.
    \item We construct \textbf{UnicBench}, a comprehensive benchmark with novel metrics to assess complex reasoning and spatial capabilities beyond simple edits. Using UnicBench, we analyze mainstream models, providing a detailed diagnosis of current limitations and a clear roadmap for future research.
\end{itemize}

%% file: tables/tab_datasets.tex
\begin{table*}[t]
\centering
\caption{Comparison of different Image Editing datasets.}
\label{tab:comp_datasets}
\setlength{\tabcolsep}{9pt}
\resizebox{\textwidth}{!}{
\renewcommand{\arraystretch}{0.9}
\begin{tabular}{l|lccccc}
\toprule[2pt] % Top thick rule (2pt)
\multirow{2}{*}{\textbf{Dataset}} & \multicolumn{1}{c}{\multirow{2}{*}{\textbf{\#Size}}} & \multirow{2}{*}{\textbf{\#Subtasks}} & \multicolumn{2}{c}{\textbf{Post Verification}} & \multirow{2}{*}{\textbf{Basic Edit}} & \multirow{2}{*}{\textbf{Complex Edit}} \\ 
\cmidrule(lr){4-5} % Rule under "Post Verification"
& & & Failed Filtration & Recaption & & \\
\midrule\midrule % Double rule under the header
MagicBrush~\cite{zhang2023magicbrush}            & $\sim$10K   & 5  & \cmark & \xmark & \cmark & \xmark \\
InstructPix2Pix~\cite{brooks2023instructpix2pix} & $\sim$313K  & 4  & \xmark & \xmark & \cmark & \xmark \\
HQ-Edit~\cite{hui2024hqedit}                     & $\sim$197K  & 6  & \xmark & \cmark & \cmark & \xmark \\
SEED-Data-Edit~\cite{ge2024seeddataedit}         & $\sim$3.7M  & 6  & \cmark & \xmark & \cmark & \xmark \\
UltraEdit~\cite{zhao2024ultraedit}               & $\sim$4M    & 9  & \xmark & \xmark & \cmark & \xmark \\
% AnyEdit             & $\sim$2.5M  & 25 & \xmark & \xmark & \cmark & \xmark \\
ImgEdit~\cite{ye2025imgedit}                     & $\sim$1.2M  & 13 & \cmark & \xmark & \cmark & \cmark \\
NHR-Edit~\cite{kuprashevich2025nohumansrequired} & $\sim$358K  & -  & \cmark & \xmark & \cmark & \xmark \\
GPT-Image-Edit-1.5M~\cite{wang2025gptimageedit}  & $\sim$1.5M  & -  & \xmark & \cmark & \cmark & \xmark \\
\midrule
\textbf{UnicEdit-10M}                                   & $\sim$10M   & 22 & \cmark & \cmark & \cmark & \cmark \\
\bottomrule[2pt] % Bottom thick rule (2pt)
\end{tabular}
}
\vspace{-0.3cm}
\end{table*}

%% file: tex/2related_works.tex
\begin{figure*}[!tb]
    \centering
    \includegraphics[width=1\textwidth]{
        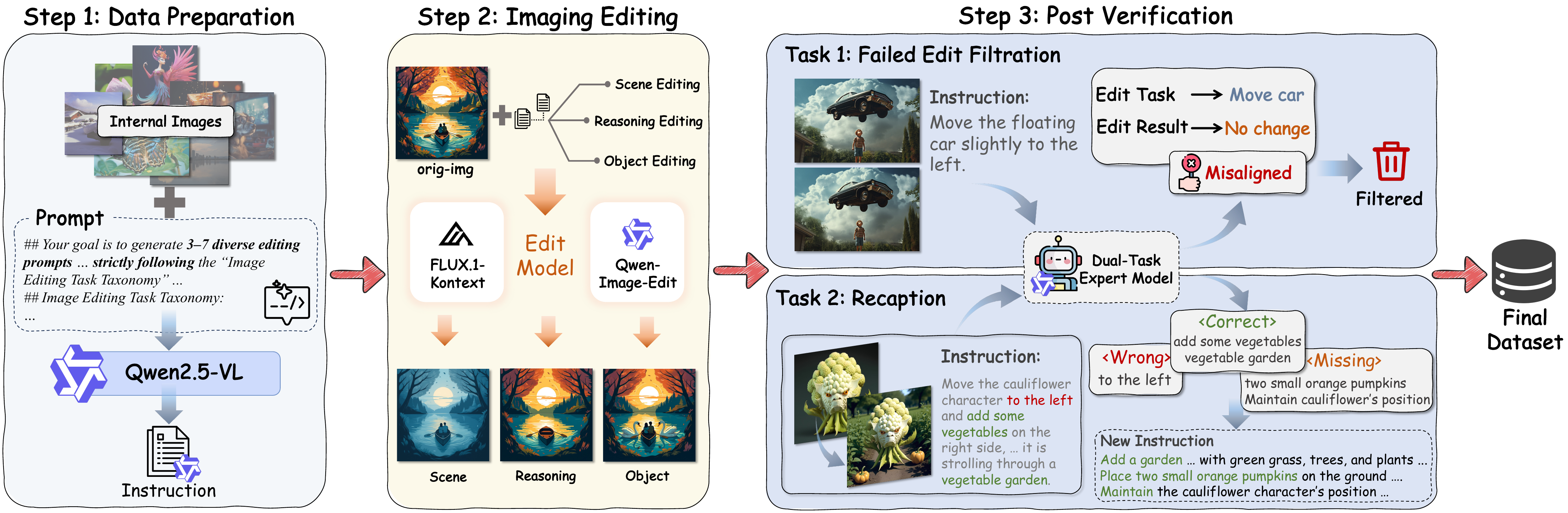
    }
    % \vspace{-0.5cm}
    \caption{Data curation pipeline with three stages: (1) data preparation, (2) image editing, (3) post verification performing failed edits filtration and recaption.}
    \vspace{-0.5cm}
    \label{fig:pipeline}
\end{figure*}
\section{Related Works}
% \paragraph{Instruction-based Image Editing.}
\noindent\textbf{Instruction-based Image Editing.}
Instruction-based image editing has progressed rapidly with diffusion models~\cite{huang2025survey-edit,shuai2024survey-edit,xu2025pgen,feng2025dit4edit}. Early methods~\cite{wang2023instructedit,li2024brushedit} often employed modular pipelines for editing, with InstructPix2Pix~\cite{brooks2023instructpix2pix} pioneering the use of synthetic data for training. Recent works like OmniGen2~\cite{wu2025omnigen2} and BAGEL~\cite{deng2025bagel} focused on unified generative and understanding abilities, while Step1X-Edit~\cite{liu2025step1x} enhanced its editing capabilities by training on a high-quality, self-curated dataset. Current open-source SOTA models, such as FLUX.1-Kontext~\cite{batifol2025flux} and Qwen-Image-Edit~\cite{wu2025qwenimageedit}, deliver high-fidelity edits through novel architectures and robust multimodal foundations. Concurrently, closed-source models like GPT-4o~\cite{hurst2024gpt4o}, Nano Banana~\cite{comanici2025nanobanana}, and Seedream 4.0~\cite{seedream2025seedream4p0} define the performance ceiling with exceptional zero-shot reasoning.

\noindent\textbf{Image Edit Datasets.}
High-quality datasets are crucial for training instruction-based editing models, with a summary in \cref{tab:comp_datasets}. Human-annotated datasets like MagicBrush~\cite{zhang2023magicbrush} offer high fidelity but limited scale. Consequently, automated pipelines are prevalent, falling into two main categories. Multi-tool pipelines, used by UltraEdit~\cite{zhao2024ultraedit}, ImgEdit~\cite{ye2025imgedit}, and Step1X-Edit~\cite{liu2025step1x}, combine various vision modules~\cite{cheng2024yolo,ravi2024sam2} but are prone to error propagation; SEED-Data-Edit~\cite{ge2024seeddataedit} mitigates this by adding human annotations. End-to-end approaches avoid error accumulation. For instance, InstructPix2Pix~\cite{brooks2023instructpix2pix} uses Stable Diffusion~\cite{rombach2022stabeldiffusion} with P2P~\cite{hertz2022prompt2prompt}, and HQ-Edit~\cite{hui2024hqedit} uses LLM~\cite{brown2020gpt3} and DALL·E 3 to generate input and output images in diptych form. However, these methods can suffer from a distribution shift due to their reliance on synthetic images, are costly, and often lack explicit quality verification. Recent works like NHR~\cite{kuprashevich2025nohumansrequired} and GPT-Image-Edit-1.5M~\cite{wang2025gptimageedit} incorporate post-filtering, but their verification is often one-dimensional and costly. In contrast, our pipeline uses open-source models with a unified post-verification process to ensure data quality in a scalable and cost-effective manner.

\noindent\textbf{Image Edit Benchmarks.}
A growing set of benchmarks targets different facets of instruction-based image editing~\cite{kawar2023imagic,huang2024smartedit}. Traditional benchmarks such as EditBench~\cite{wang2023editbench} focus on specific tasks like inpainting, while EmuEdit~\cite{sheynin2024emuedit} and AnyEdit~\cite{yu2025anyedit} rely on metrics like L1 and CLIP~\cite{radford2021clip} score, which may misalign with human perception of edit faithfulness. Recent works like ImgEdit-Bench~\cite{ye2025imgedit} and GEdit-Bench~\cite{liu2025step1x} span diverse tasks and leverage VLMs for assessment, yet are constrained by few complex cases or metrics that are insensitive to unintended changes in non-edited regions. KRIS-Bench~\cite{wu2025krisbench} specifically probes complex reasoning but omits basic edits, limiting its role as a general benchmark. Our benchmark addresses these gaps with comprehensive coverage of both basic and complex edits and introduces fine-grained metrics for unintended-change control and reasoning fidelity, enabling more diagnostic evaluation.

%% file: tex/3methodology.tex
\begin{figure*}[t]
    \centering
    \includegraphics[width=1\textwidth]{
        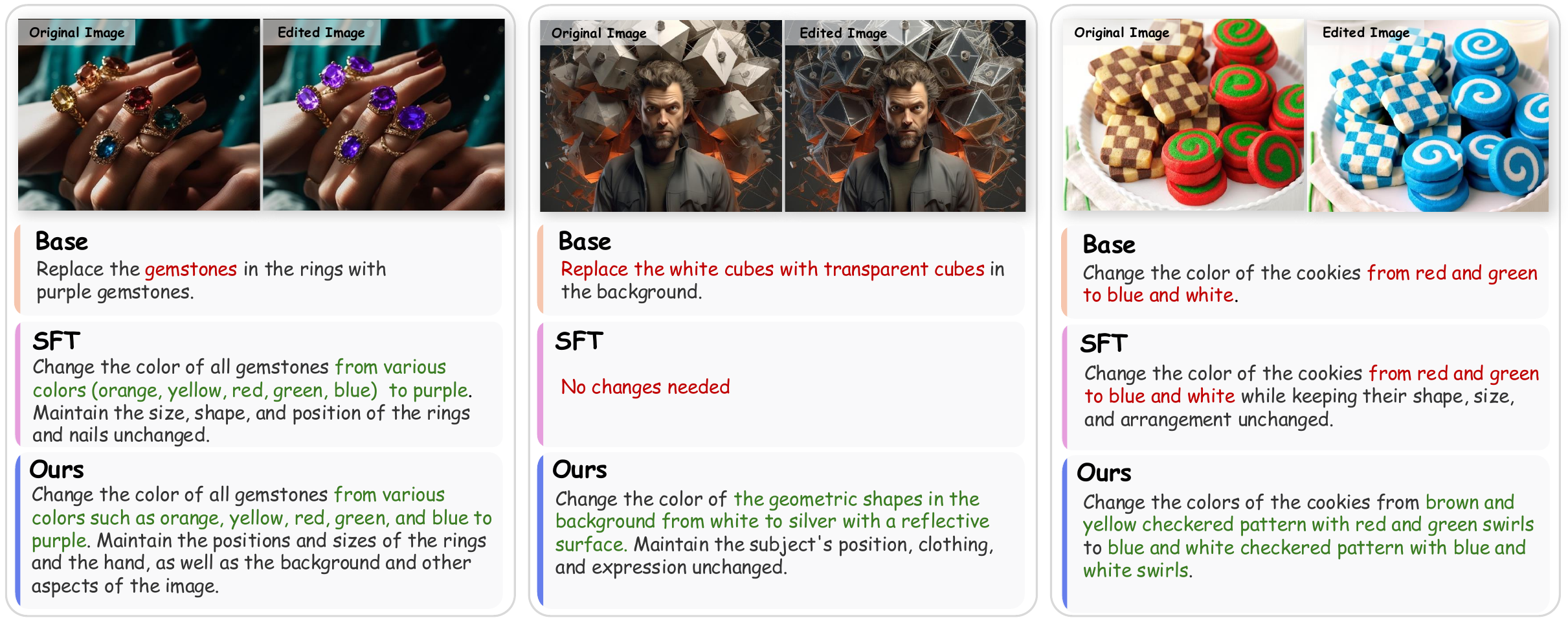
    }
    % \vspace{-0.5cm}
    \caption{Post-verification examples of the expert model. \textit{Base} denotes Qwen2.5-VL-7B; \textit{SFT} denotes Base model after Stage-1 SFT; \textit{Ours} denotes the dual-task expert model \textbf{Qwen-Verify}.}
    \vspace{-0.5cm}
    \label{fig:cases_captioner}
\end{figure*}

\section{UnicEdit-10M Dataset}
% Our methodology centers on a scalable pipeline for creating a large-scale, high-fidelity image editing dataset. This section details our data generation pipeline (\cref{sec3p1:pipeline}) and the expert model trained for post-verification (\cref{sec3p2:expert_model}).
In this section, we propose a scalable data generation pipeline to curate UnicEdit-10M, a large-scale, high-quality image editing dataset.

\subsection{Dataset Preparation}
\noindent\textbf{Editing Taxonomy.} 
As shown in \cref{fig:teaser}, to ensure comprehensive task coverage, we established a fine-grained taxonomy that classifies edits into four high-level categories: \textit{Object Editing}, \textit{Attribute Editing}, \textit{Scene Editing}, and \textit{Reasoning Editing}, totaling 22 sub-tasks. We define tasks requiring spatial awareness and factual knowledge as Complex Edits, with all others classified as Basic Edits. This two-level classification ensures a comprehensive task coverage. Further details are in the supplementary material (\cref{suppl:pipeline_taxonomy}).

\noindent\textbf{Data Source.}
Our pipeline begins with a large-scale internal library of diverse \textit{real-world} and \textit{synthetic} images, all pre-filtered for high aesthetic scores to ensure quality.

\noindent\textbf{Model Selection.}
For instruction generation, we selected Qwen2.5-VL-72B~\cite{bai2025qwen2p5vl} over its 7B counterpart due to its superior visual comprehension and ability to generate accurate and diverse instructions. For the image editing module, we employed FLUX.1-Kontext~\cite{batifol2025flux} and Qwen-Image-Edit~\cite{wu2025qwenimageedit}, two leading open-source models renowned for their proficiency across a broad spectrum of editing tasks.

\subsection{Dataset Curation}
\label{sec3:dataset_curation}
Our pipeline is a three-stage process: instruction generation, image editing, and a crucial post-verification step, as shown in \cref{fig:pipeline}.
% \paragraph{Instruction Generation.}

\noindent\textbf{Instruction Generation.}
We automate instruction generation by prompting Qwen2.5-VL-72B~\cite{bai2025qwen2p5vl} with a source image and predefined editing taxonomy. The model generates 3-7 distinct, content-aware instructions per image, each corresponding to a unique sub-task, 
ensuring a balanced task distribution without manual annotation.

\noindent\textbf{Image Editing.}
Each ⟨\textit{original image, instruction}⟩ pair is then processed by editing models to synthesize an edited image, forming an initial triplet. To handle varying input resolutions, all source images are preprocessed via center-cropping and resizing, with a quality check to discard images requiring more than 20\% cropping to prevent significant content loss. However, this automated process is not infallible. Failures arise from ambiguous instructions (\eg, ``remove the bag from the person" in a multi-person scene) or inherent limitations in the editing model's semantic grounding. This can result in unchanged images or unexpected edit, necessitating a rigorous post-verification stage.

\noindent\textbf{Post-Verification and Refinement.}
The raw generated triplets contain noisy data that can impair model training. We identify three primary failure modes:
\begin{itemize}
    \item \textit{Edit Failure}: The edited image is identical or nearly identical to the source.
    \item \textit{Instruction-Image Misalignment}: The model performs unintended edits or fails to follow the instruction accurately.
    \item \textit{Others}: Less frequent but critical failures, such as degraded quality or anatomical errors, detailed in the supplementary material (\cref{suppl:analy_noise}).
\end{itemize}
To overcome these limitations, we introduce an expert verification framework that integrates failure filtration and instruction recaptioning into a unified process driven by Chain-of-Thought~\cite{wei2022cot} reasoning. We employ Qwen2.5-VL-72B~\cite{bai2025qwen2p5vl} as our expert verifier, guided by a multi-step prompt. The process unfolds as follows: First, the model captions the original and edited images to expose fine-grained changes. Second, it decides whether a valid edit occurred from their visual differences. Then, for valid edits, the model rewrites the instruction to precisely match the visual modification. Finally, it outputs a structured JSON object containing: (1) a boolean flag, \textit{is\_changed}, indicating a valid edit, and (2) the refined instruction. However, this approach is computationally expensive, and the recaptioned instructions are \textit{prone to hallucination}. To address these issues, we developed a specialized expert model for this task, which we detail in \cref{sec4:expert_model}. This expert model is ultimately deployed in our pipeline for large-scale data curation.

% This CoT-driven methodology enhances the verifier's attention to detail and logical consistency, drastically reducing misclassifications.

\subsection{Dataset Statistics}
Our pipeline produces a high-quality dataset comprising approximately 10M triplets, organized into 22 fine-grained editing categories. The dataset is structurally divided into four major editing types: Scene Editing (3.063M samples), Attribute Editing (3.529M), Object Editing (3.242M), and Reasoning Editing (1.746M). In terms of visual quality, the dataset predominantly consists of high-resolution images, with 1024×1024 resolution accounting for 50\% of the total collection. More details can be found in the the supplementary material (\cref{suppl:dataset_comparison}).
% Recent instruction-based image editing models have demonstrated remarkable progress in handling complex user commands, yet widely adopted benchmarks such as GEdit and ImgEdit remain limited in their coverage of challenging editing scenarios. This omission prevents a systematic assessment of model capabilities across the full spectrum of editing difficulty—from basic object-level changes to semantically or geometrically demanding transformations.

\input{tables/tab_dataset_vie}
\section{Qwen-Verify}
\label{sec4:expert_model}
While leveraging a large model like Qwen2.5-VL-72B~\cite{bai2025qwen2p5vl} for post-verification is effective, it is computationally expensive and still \textit{prone to hallucination}. To address these issues, we introduce Qwen-Verify, a 7B expert model trained to perform our dual-task verification with high efficiency. 

\subsection{Data}
Our training data targets dominant failure modes and is partitioned into three categories. All samples are meticulously screened and corrected by human annotators to ensure high fidelity.
\begin{itemize}
    \item \textit{Normal}: High-quality triplets with salient, instruction-aligned edits.
    \item \textit{No Edit}: Triplets with no discernible change between the original and edited images.
    \item \textit{Hallucination}: Cases where the target is correct but the action or attribute is misstated.  % We derive these via recaptioning during post-verification and then finalize the corrected instructions through human verification.
\end{itemize}
The training data is assembled in stages: 
First, we prepare 200k samples for SFT, consisting of \textit{Normal} and \textit{No Edit} data. \textit{Normal} data instructions are reviewed and refined by human experts. \textit{No Edit} aggregates (1) triplets rejected during post-verification and (2) no-op triplets that slipped through automated screening but were later flagged by annotators.
Next, We use all three categories to construct 20k DPO samples. \textit{Normal} and \textit{No Edit} follow the same construction as SFT. For the \textit{Hallucination} set, we select post-verified triplets with recaptioned instructions and then correct them to create high-quality preference pairs. This correction process involves using GPT-4o~\cite{hurst2024gpt4o} to generate candidate corrections, which are then rigorously reviewed and finalized by human experts before DPO training.

\subsection{\textbf{\DPO}}
We developed Qwen-Verify by training the Qwen2.5-VL-7B~\cite{bai2025qwen2p5vl} in a two-stage training strategy: supervised fine-tuning (SFT) for foundational capabilities, followed by a preference alignment stage~\cite{rafailov2023dpo} to refine the model's understanding of nuanced instruction quality.

We first perform SFT on the base model using a mix of \textit{Normal} and \textit{No Edit} data. This stage equips the model with the core abilities to distinguish failed edits and generate accurate instructions for successful ones.

While SFT provides a strong starting point, it is agnostic to the subtle yet critical semantic differences between adequate and excellent instructions. To address this, in the second stage, we introduce \textbf{Differential Direct Preference Optimization (\DPO)}, a novel framework that adapts preference optimization for the challenge of interpreting visual changes.
Unlike conventional methods that treat visual inputs as static context, \DPO\ re-frames the learning problem by conditioning the policy on a dynamically computed \textit{visual differential context}. We define a visual encoder, $\mathcal{V}$, that processes the image pair to extract a latent representation of the edit:
\begin{equation}
    c_{v}= \mathcal{V}(I_{o}, I_{e})
\end{equation}
This vector $c_{v}$ encapsulates the transformation from $I_{o}$ to $I_{e}$ and conditions optimization. We learn from a curated preference set $\mathcal{D}= \{(I_{o}, I_{e}, p_{w}, p_{l})\}_{i}$, where $(p_{w}, p_{l})$ are preferred (corrected) and rejected (hallucination) instructions. Assuming a latent reward $r(p, c_{v})$, the Bradley-Terry model gives $P(p_{w}\succ p_{l}| c_{v}) = \sigma(r(p_{w}, c_{v}) - r(p_{l}, c_{v}))$. \DPO\ avoids explicit reward modeling by re-parameterizing rewards via policy probabilities. We define the Policy Advantage Function:
\begin{equation}
    \mathcal{A}_{\pi_\theta, \pi_{\text{ref}}}(p, c_{v}) = \beta \log \frac{\pi_{\theta}(p
    | c_{v})}{\pi_{\text{ref}}(p | c_{v})}
\end{equation}
where $\pi_{\theta}$ is the trainable policy, $\pi_{\text{ref}}$ is a frozen SFT copy, and $\beta$ controls deviation. The objective maximizes the advantage margin:
\begin{equation}
    \label{eq:d2po_advanced}
    \begin{aligned}
        \mathcal{L}_{\text{\DPO}} & = -\mathbb{E}_{(c_v, p_w, p_l) \sim \mathcal{D}}                                                                                                              \\
                                  & \left[ \log \sigma \left( \mathcal{A}_{\pi_\theta, \pi_{\text{ref}}}(p_{w}, c_{v}) - \mathcal{A}_{\pi_\theta, \pi_{\text{ref}}}(p_{l}, c_{v}) \right) \right]
    \end{aligned}
\end{equation}
Optimizing this loss aligns the model with human judgments of precise and faithful edit descriptions.

\section{UnicBench}

To enable a more comprehensive assessment of model capabilities, we introduce UnicBench, a unified benchmark spanning both basic and complex edits. It is accompanied by fine-grained metrics designed to evaluate model performance across these different dimensions.

\subsection{Benchmark Construction}
UnicBench is built upon a diverse set of high-quality \textit{real-world} and \textit{synthetic} images. We generate instructions using a hybrid VLM-human workflow: Qwen2.5-VL~\cite{bai2025qwen2p5vl} first produces a set of candidate instructions based on image content, and then we employ human experts to review the resulting ⟨image, instruction⟩ pairs. This step removes ambiguous or semantically inconsistent prompts, with subsequent rewriting applied to align instructions with specific editing task categories. The benchmark follows the same taxonomy as the training data, with 50 test cases per category. As detailed in the supplementary material (\cref{suppl:comp_bench}), a comparative analysis confirms that UnicBench offers more comprehensive task coverage than existing benchmarks, enabling a more precise and multidimensional assessment of model capabilities.

\subsection{Evaluation Metric}
While the VIEScore~\cite{ku2023viescore} provides a foundational assessment, it suffers from key limitations. Its Semantic Consistency (SC) metric is insensitive to unintended alterations in non-edited regions, and its Perceptual Quality (PQ) metric prioritizes naturalness, potentially underestimating stylized outputs. Crucially, neither metric effectively handles reasoning-based nor geometrically complex edits. To address these gaps, we propose a new evaluation framework composed of four specialized metrics:
\begin{itemize}
    \item \textit{Instruction Following (IF):} Measures how well the edited image satisfies the instruction via a VLM-based cross-modal alignment score.
    \item \textit{Non-edit Consistency (NC):} Assesses preservation of non-target regions, penalizing unintended changes outside the specified edit area.
    \item \textit{Visual Quality (VQ):} Instruction-conditioned assessment of naturalness, coherence, and adherence to the intended visual style.
    \item \textit{Reasoning Accuracy (RA):} Targets knowledge-intensive edits. A VLM first derives an intended-outcome specification from the instruction and original image. To ground this inference, each sample provides a \textit{list of reasoning-points} (targets, operations, expected visual changes), which guides the verifier’s attention. The edited image is then checked against this specification.
\end{itemize}
Each metric is scaled from 0 to 10, and a score of zero in any category indicates a failed edit. The overall edit score is computed as the geometric mean of all applicable metrics for a given task. This is formalized as:
\begin{equation}
    \text{Score} = \left( \prod_{m \in \mathcal{M}} m \right)^{1/|\mathcal{M}|}
\end{equation}
where $\mathcal{M}$ is the set of relevant metrics for the task category: $\mathcal{M} = \{IF, NC, VQ\}$ for Basic Edits, and $\mathcal{M} = \{IF, NC, VQ, RA\}$ for Complex Edits. This formulation ensures that severe failures in any dimension are reflected in the final score, providing a more reliable and interpretable measure of editing performance. 
% We provide a comparative analysis of VIE and our metric in the supplementary material (\cref{suppl:analy_metric}).

%% file: tables/tab_dataset_vie.tex
\begin{table*}[ht]
\centering
\caption{Overall dataset quality comparison between \textbf{UnicEdit-10M} and other datasets. VIEScore reports Semantic Consistency (SC) and Perceptual Quality (PQ) with their Overall score. Aesthetics columns give the aesthetic scores for the Source (original) and Target (edited) images. Best results are in \textbf{bold}, and second best are \underline{underlined}.}
\label{tab:res_datasets}
% Using resizebox to maintain style and ensure it fits within page boundaries
\setlength{\tabcolsep}{28pt}
\resizebox{\textwidth}{!}{
\renewcommand{\arraystretch}{0.9}
\begin{tabular}{l|ccc|cc}
\toprule[2pt] % Top thick rule (2pt)
\multirow{2}{*}{\textbf{Datasets}} & \multicolumn{3}{c|}{\textbf{VIEScore $\uparrow$}} & \multicolumn{2}{c}{\textbf{Aesthetics $\uparrow$}} \\
\cmidrule(lr){2-4} \cmidrule(lr){5-6} % Rules under the top-level headers
& \textbf{SC} & \textbf{PQ} & \textbf{Overall} & \textbf{Source} & \textbf{Target} \\
\midrule\midrule % Double rule under the header, as per your reference style
SEED-Data-Edit~\cite{ge2024seeddataedit}                 & 5.7884 & 6.3430 & 5.0043 & 5.72 & 5.74 \\
ImgEdit~\cite{ye2025imgedit}                             & 6.3233 & 7.8819 & 6.2462 & 6.49 & 7.03 \\
X2Edit~\cite{ma2025x2edit}                               & 7.3527 & 7.2776 & 6.8693 & 7.52 & 7.54 \\
NHR-Edit~\cite{kuprashevich2025nohumansrequired}  & 8.3180 & 7.9420 & \underline{7.7796} & \underline{7.35} & 7.42 \\
GPT-Image-Edit-1.5M~\cite{wang2025gptimageedit}          & \textbf{8.6780} & 7.1560 & 7.7451 & 6.23 & \underline{7.59} \\
Nano-consistent-150k~\cite{ye2025nanoconsistent}         & 7.9180 & \underline{8.0000} & 7.7520 & 6.81 & 7.40 \\
\midrule % Single rule before the "Ours" row, as per your reference style
\textbf{Ours}                                           & \underline{8.4500} & \textbf{8.1950} & \textbf{8.0768} & \textbf{8.00} & \textbf{7.76} \\
\bottomrule[2pt] % Bottom thick rule (2pt)
\end{tabular}
}
\vspace{-0.3cm}
\end{table*}

% caption解释清楚

%% file: tex/4experiments.tex
\section{Experiments}
\subsection{Evaluation Models \& Settings}
For Dataset Quality, we employ the \textit{VIEScore}~\cite{ku2023viescore} to compute both the Semantic Consistency (SC) and Perceptual Quality (PQ) for mainstream datasets, and employ \textit{Aesthetic score} for image aesthetics of source and edited images.
For Expert Model, we use Qwen2.5-VL-7B and Qwen2.5-VL-72B~\cite{bai2025qwen2p5vl} as baselines.
For Benchmark Evaluation, we test some representative models on UnicBench. These include the closed-source Nano Banana~\cite{comanici2025nanobanana}, SeedEdit 3.0~\cite{wang2025seededit3}, Seedream 4.0~\cite{seedream2025seedream4p0}, and GPT-4o~\cite{hurst2024gpt4o}, alongside open-source models such as Instruct-Pix2Pix~\cite{brooks2023instructpix2pix}, MagicBrush~\cite{zhang2023magicbrush}, UniWorld-V1~\cite{lin2025uniworldv1}, OmniGen2~\cite{wu2025omnigen2}, BAGEL~\cite{deng2025bagel}, NextStep-1~\cite{nextstepteam2025nextstep1}, Step1X-Edit-v1.1~\cite{liu2025step1x}, FLUX.1-Kontext~\cite{batifol2025flux}, and Qwen-Image-Edit~\cite{wu2025qwenimageedit}. Models that support Chinese are evaluated with both English and Chinese prompts. All assessments are conducted by \textit{gpt-4.1-2025-04-14} for consistent scoring.

\input{tables/tab_dataset_quantity}

\subsection{UnicEdit-10M Dataset Evaluation}
% \subsubsection{Dataset Analysis}
\noindent\textbf{Quantitative.}
The data volume at each pipeline stage is detailed in \cref{tab:data_quantity}. For each source image, we generate multiple candidate prompts and select 3–5 to produce edited images. Notably, post-verification with Qwen-Verify filters 26\% failed edits, ensuring the final set contains valid image pairs.
For quantitative comparison (\cref{tab:res_datasets}), we adopt the X2Edit~\cite{ma2025x2edit} protocol, employing GPT-4o~\cite{hurst2024gpt4o} to assesse 1K randomly sampled triplets, averaged over three runs. UnicEdit-10M attains the highest PQ and overall score, and surpasses all competitors by a large margin in Aesthetic score, validating our post-verification process. While both our dataset and GPT-Image-Edit-1.5M~\cite{wang2025gptimageedit} achieve high SC due to the instruction recaptioning step in both pipelines, a targeted analysis of face consistency reveals a key difference: UnicEdit achieves a robust consistency score of \textit{0.89}, significantly outperforming GPT-Image-Edit-1.5M's \textit{0.3025}. This highlights our method's superior ability to maintain critical subject details, a crucial factor for training reliable editing models. 

\noindent\textbf{Qualitative.}
Qualitative samples of UnicEdit-10M are shown in \cref{fig:showcases}. We also conduct targeted case studies to visually demonstrate our pipeline's effectiveness, as shown in the supplementary material (\cref{suppl:b1_construction_method,suppl:facial}). We process samples from ImgEdit and SEED-Data-Edit, which initially suffer from low edit quality and instruction misalignment. Our pipeline significantly enhances these triplets, correcting errors and improving semantic alignment. Furthermore, a comparison of facial consistency reveals that UnicEdit-10M is substantially better at preserving key facial features than GPT-Image-Edit-1.5M, validating our method's ability to produce high-fidelity training data.

\input{tables/tab_benchmark_result}
\input{tables/tab_captioner}

% \subsubsection{Expert Model Analysis}
\subsection{Expert Model Performance}
\noindent\textbf{Qwen-Verify vs. Qwen2.5-VL-72B.}
To evaluate Qwen-Verify, we curated a test set spanning the three defined tasks and measured performance via \textit{Alignment Accuracy (Acc.)}, calculated by GPT-4.1. More details are in the supplementary material (\cref{suppl:expert_eval}). As shown in \cref{tab:res_captioner}, our model significantly outperforms all baselines, including the strong Qwen2.5-VL-72B, across all test categories. Qualitative cases (\cref{fig:cases_captioner}) further show that Qwen-Verify captures fine-grained visual changes and produces instructions precisely aligned with the edits. This strength stems from its dual-task design, which jointly optimizes failure detection and instruction refinement, enabling it to grasp subtle semantic nuances.

\noindent\textbf{Qwen-Verify vs. SSIM.}
We also compare Qwen-Verify against the traditional SSIM method for filtering \textit{No edit} pairs, as discussed in the supplementary material (\cref{suppl:b2_no_edit}). Qwen-Verify achieves significantly higher accuracy. Case studies reveal that SSIM is insensitive to semantically meaningful but visually subtle changes, while being overly sensitive to the minor, imperceptible pixel shifts inherent in the generation process. This confirms that traditional vision-based methods are ill-suited for our semantically-driven post-verification task.

\subsection{UnicBench Analysis}
\noindent\textbf{Model Performance on UnicBench.}
We present a comprehensive evaluation of mainstream models on UnicBench in \cref{tab:res_bench_overall}. The results confirm that closed-source models exhibit more powerful and comprehensive capabilities, consistently outperforming open-source models. GPT-Image-1~\cite{hurst2024gpt4o} achieves the highest overall score on both EN and CN tasks, demonstrating SOTA general editing ability. Seedream 4.0~\cite{seedream2025seedream4p0} follows as the second-best model, even surpassing GPT-Image-1~\cite{hurst2024gpt4o} on the NC metric. Among open-source models, Qwen-Image-Edit~\cite{wu2025qwenimageedit} leads in performance, beginning to close the gap. A key insight from the metric breakdown is that while most models effectively follow basic instructions, nearly all exhibit a significant performance drop on the RA metric. This reveals a common limitation in executing edits that require complex logical inference or world knowledge, suggesting a clear direction for future research. This underscores the importance of our work, which provides a high-quality dataset to train more advanced models, while offers a scalable method for generating specific task data.

\noindent\textbf{Metric Comparison with VIEScore.}
We compare our evaluation metrics against VIEScore, with detailed cases in the supplementary material (\cref{suppl:analy_metric}). Our analysis reveals that while VIEScore can accurately assess whether an instruction was executed, it is largely insensitive to unintended changes in non-edited regions. For instance, in cases involving the unwanted removal of a person or accidental modification of text, VIEScore still assigned a high SC score. In contrast, our metrics decouple this assessment into Instruction Following (IF) and Non-edit Consistency (NC). In the same failure cases, our NC dimension correctly identified the unintended changes and penalized the score accordingly. This comparison demonstrates that our metrics provide a more fine-grained and diagnostic evaluation, enabling a precise localization of specific model deficiencies.

%% file: tables/tab_dataset_quantity.tex
\begin{table}[ht]
\centering
\caption{Data volume statistics at each stage of our pipeline. Ratio indicates the percentage change from the previous stage, and Gen. abbreviates Generation. FLUX and Qwen refer to FLUX.1-Kontext~\cite{batifol2025flux} and Qwen-Image-Edit~\cite{wu2025qwenimageedit}, respectively.}
\label{tab:data_quantity}
% Use resizebox to maintain style and fit.
\resizebox{\columnwidth}{!}{
\begin{tabular}{l|lll}
\toprule[2pt] % Top thick rule (2pt)
\textbf{Processing Stage} & \textbf{Method} & \textbf{$\Delta$ Ratio(\%)} & \textbf{Data Volume} \\
\midrule\midrule % Double rule under the header, as per your reference style
Initial Images          & Internal Images   & -         & 5001199 \\
Instruction Gen.  & Qwen2.5-VL-72B    & +447.26   & 22368563 \\
Editing Gen.      & FLUX/Qwen         & -30.03    & 15651530 \\
Failed Edit Filter  & Qwen-Verify       & -25.97    & 11586583 \\
Recaption               & Qwen-Verify       & -         & 11586583 \\
\midrule % Single rule before the "Final Data" row
\textbf{Final Data}     & -                 & -         & \textbf{11586583} \\
\bottomrule[2pt] % Bottom thick rule (2pt)
\end{tabular}
}
\vspace{-0.3cm}
\end{table}

%% file: tables/tab_benchmark_result.tex
\begin{table*}[ht]
\centering
\caption{Overall performance of different model on UnicBench. The performance of open-source and closed-source models is separately marked with the best performance in \textbf{bold}, and the second best \underline{underlined}.}
% \vspace{-0.3cm}
\label{tab:res_bench_overall}
% Use resizebox to ensure the table fits within the text width
\setlength{\tabcolsep}{9pt}
\resizebox{\textwidth}{!}{
\renewcommand{\arraystretch}{0.95}
\begin{tabular}{l|ccccc|ccccc}
\toprule[2pt] % Top thick rule (2pt)
\multirow{2}{*}{\textbf{Model}} & \multicolumn{5}{c|}{\textbf{Overall-EN}} & \multicolumn{5}{c}{\textbf{Overall-CN}} \\
\cmidrule(lr){2-6} \cmidrule(lr){7-11} % Rules under the top-level headers
& \textbf{IF} & \textbf{NC} & \textbf{VQ} & \textbf{RA} & \textbf{Overall} & \textbf{IF} & \textbf{NC} & \textbf{VQ} & \textbf{RA} & \textbf{Overall} \\
\midrule\midrule % Double rule under the header, as per your reference style
\rowcolor{gray!20} % Set background color for this row
\multicolumn{11}{c}{\textit{Open-source Models}} \\ % Sub-header for the group
Instruct-Pix2Pix~\cite{brooks2023instructpix2pix}    & 2.8526 & 4.0983 & 3.9672 & 1.9560 & 2.9221 & - & - & - & - & - \\
MagicBrush~\cite{zhang2023magicbrush}                & 2.3403 & 3.3849 & 3.4559 & 1.7240 & 2.3407 & - & - & - & - & - \\
OmniGen2~\cite{wu2025omnigen2}                       & 6.2455 & 7.4973 & 6.4891 & 5.1240 & 6.1246 & - & - & - & - & - \\
UniWorld-v1~\cite{lin2025uniworldv1}                 & 5.3055 & 7.3091 & 6.4827 & 4.0160 & 5.6013 & - & - & - & - & - \\
FLUX.1-Kontext~\cite{batifol2025flux}                & 6.7755 & \textbf{8.4718} & {\ul 7.3600} & 5.5040 & 6.8045 & - & - & - & - & - \\
NextStep-1~\cite{nextstepteam2025nextstep1}          & 7.2311 & 6.1802 & 6.5014 & {\ul 6.2920} & 6.4076 & - & - & - & - & - \\
BAGEL~\cite{deng2025bagel}                           & {\ul 7.2491} & 8.1982 & 7.1391 & 5.2600 & {\ul 6.9794} & {\ul 7.3018} & {\ul 8.2845} & 7.3118 & {\ul 5.2840} & {\ul 7.1056} \\
Step1X-Edit-v1.1~\cite{liu2025step1x}                & 6.9945 & {\ul 8.2045} & 7.3382 & 5.0400 & 6.9202 & 7.0282 & \textbf{8.4118} & {\ul 7.5600} & 5.0560 & 7.0620 \\
Qwen-Image-Edit~\cite{wu2025qwenimageedit}           & \textbf{8.2055} & 8.0264 & \textbf{8.0745} & \textbf{6.4480} & \textbf{7.7273} & \textbf{8.3718} & 7.8000 & \textbf{8.2118} & \textbf{6.6560} & \textbf{7.7790} \\
\midrule % Rule to separate the two groups
\rowcolor{gray!20} % Set background color for this row
\multicolumn{11}{c}{\textit{Closed-source Models}} \\ % Sub-header for the group
Nano Banana~\cite{comanici2025nanobanana}            & 7.9753 & \textbf{8.9808} & {\ul 8.1954} & 6.8680 & 7.8792 & 8.1550 & \textbf{9.0438} & {\ul 8.3291} & 6.8960 & 8.0358 \\
SeedEdit 3.0~\cite{wang2025seededit3}                & 8.2717 & 8.4251 & 7.8392 & 6.9393 & 7.8671 & {\ul 8.3721} & 8.4502 & 7.9795 & 6.8395 & 7.9753 \\
Seedream 4.0~\cite{seedream2025seedream4p0}          & {\ul 8.3764} & {\ul 8.7200} & 8.0736 & {\ul 7.5960} & {\ul 8.0428} & 8.3418 & {\ul 8.6600} & 8.1364 & {\ul 7.1240} & {\ul 8.0474} \\
GPT-Image-1~\cite{hurst2024gpt4o}                    & \textbf{9.1551} & 7.8449 & \textbf{8.6830} & \textbf{8.3392} & \textbf{8.3546} & \textbf{9.2759} & 7.8906 & \textbf{8.6980} & \textbf{8.2247} & \textbf{8.4506} \\
\bottomrule[2pt] % Bottom thick rule (2pt)
\end{tabular}
}
\vspace{-0.3cm}
\end{table*}

%% file: tables/tab_captioner.tex
% \begin{table}[ht]
% \centering
% \caption{Performance of post-verification expert model.}
% \label{tab:res_captioner}
% \setlength{\tabcolsep}{3pt}
% \resizebox{\columnwidth}{!}{
% \begin{tabular}{l|ccc}
% \toprule[2pt] % Top thick rule (2pt)
% \textbf{Model} & \textbf{Normal Acc. $\uparrow$} & \textbf{No Edit Acc. $\uparrow$} & \textbf{Hallucination Acc. $\uparrow$} \\
% \midrule\midrule % Double rule under the header, as per your reference style
% Qwen2.5-VL-7B        & 4.39 & 4.84 & 3.95 \\
% Qwen2.5-VL-72B       & 5.25 & 9.60 & 6.12 \\
% Qwen2.5-VL-7B+SFT    & 5.62 & 9.40 & 5.47 \\ % Indented with \quad
% \midrule
% \textbf{Qwen-Verify} & \textbf{6.32} & \textbf{9.80} & \textbf{6.22} \\ % Indented with \quad
% \bottomrule[2pt] % Bottom thick rule (2pt)
% \end{tabular}
% }
% \vspace{-0.3cm}
% \end{table}

\begin{table}[ht]
\centering
\caption{Performance of post-verification expert model.}
\label{tab:res_captioner}
\setlength{\tabcolsep}{9pt}
\resizebox{\columnwidth}{!}{
\begin{tabular}{l|ccc}
\toprule[2pt]
\textbf{Model} & \textbf{\makecell[c]{Normal \\ Acc. $\uparrow$}} & \textbf{\makecell[c]{No Edit \\ Acc. $\uparrow$}} & \textbf{\makecell[c]{Hallucination \\ Acc. $\uparrow$}} \\
\midrule\midrule
Qwen2.5-VL-7B         & 4.39 & 4.84 & 3.95 \\
Qwen2.5-VL-72B       & 5.25 & 9.60 & 6.12 \\
Qwen2.5-VL-7B+SFT    & 5.62 & 9.40 & 5.47 \\ 
\midrule
\textbf{Qwen-Verify} & \textbf{6.32} & \textbf{9.80} & \textbf{6.22} \\ 
\bottomrule[2pt]
\end{tabular}
}
\vspace{-0.3cm}
\end{table}

%% file: tex/5conclusion.tex
\section{Conclusion}
We introduce a scalable pipeline featuring the Qwen-Verify expert model for data curation, yielding UnicEdit-10M, a 10M dataset covering 22 editing tasks. We also presented UnicBench, a comprehensive benchmark with novel metrics for fine-grained diagnosis. Our analysis on UnicBench pinpoints complex reasoning and spatial understanding as the critical bottleneck for current models. Our scalable pipeline provides a direct path forward, enabling the targeted, automated generation of large-scale training data for these specific tasks. By releasing the dataset, benchmark, and reusable pipeline, we aim to provide the community with essential resources to accelerate progress and close the capability gap with closed-source models.

%% file: tex/9suppl.tex
\clearpage
\setcounter{page}{1}
\maketitlesupplementary

\appendix

\begin{strip}
  \centering
  \includegraphics[width=\textwidth]{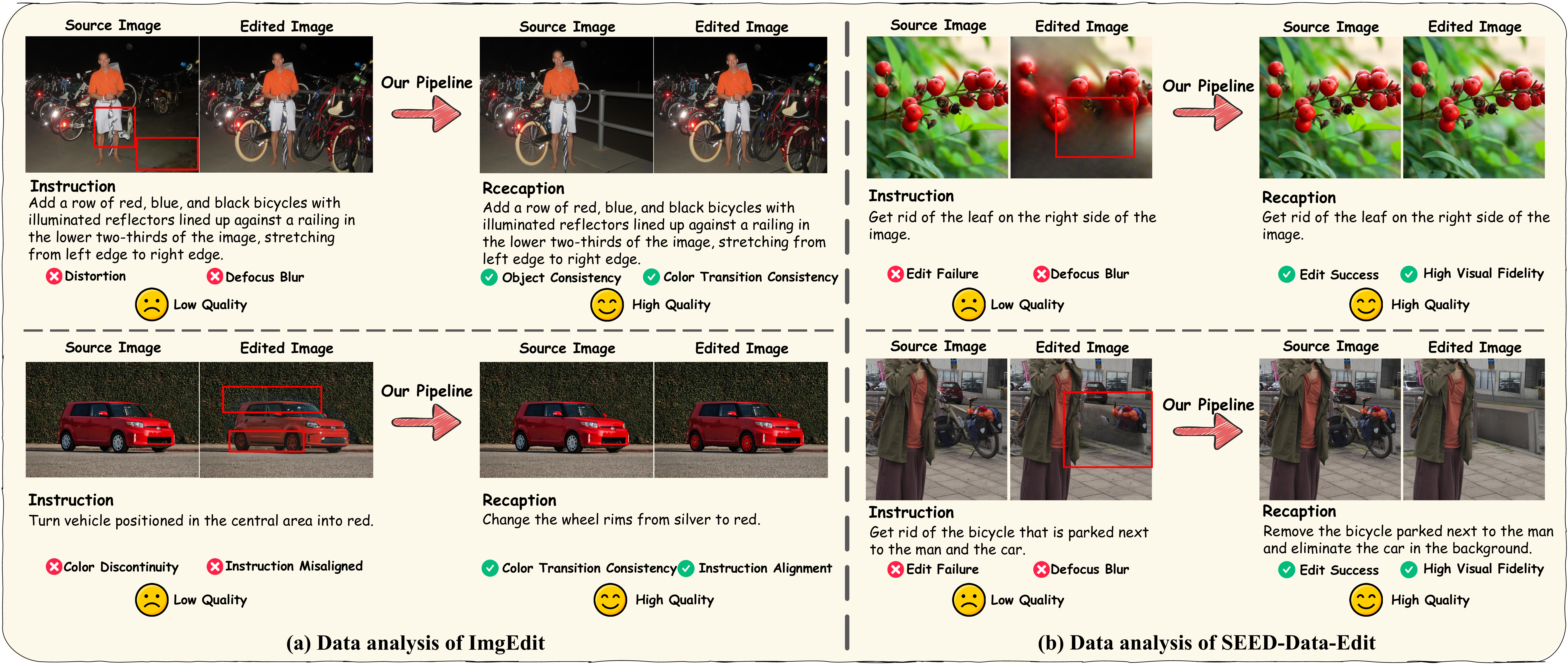}
  \captionof{figure}{Qualitative comparison of data curation pipelines. (a) shows example triplets from ImgEdit~\cite{ye2025imgedit}. (b) shows example from SEED-Data-Edit~\cite{ge2024seeddataedit}. In each subfigure, the left columns display original triplets, while the right columns show the same images reprocessed through \textit{Our Pipeline}. \red{Red boxes} highlight regions of blurring, artifacts, and color inconsistencies in the originals. Our pipeline consistently yields higher quality and more precisely aligned instructions, demonstrating the effectiveness of our unified verification process.}
  \label{fig:cases_dataset_sample}
\end{strip}

% \begin{figure*}[!tb]
%     \centering
%     \includegraphics[width=1\textwidth]{figures/cases_datasetsamples_v1.pdf}
%     % \vspace{-0.5cm}
%     \caption{dataset pipeline}
%     % \vspace{-0.6cm}
%     \label{fig:cases_dataset_sample}
% \end{figure*}

In this supplementary material, we further elaborate on three core components of our framework: the data construction pipeline, the expert verification model, and the UnicBench benchmark. For the \textbf{Data Construction Pipeline}, we present the dataset taxonomy in \cref{suppl:pipeline_taxonomy}, analyze our pipeline and post-verification procedures in \cref{suppl:pipeline_analysis}, discuss noisy cases in \cref{suppl:analy_noise}, and study facial consistency in \cref{suppl:facial}. For the \textbf{Expert Model Evaluation}, we describe the experimental setup and metrics used to assess our verification model in \cref{suppl:expert_eval}. For the \textbf{Benchmark Analysis}, we validate the evaluation metrics in \cref{suppl:analy_metric}, report comprehensive benchmark results in \cref{suppl:bench_res}, compare UnicBench with existing benchmarks in \cref{suppl:comp_bench}, and provide qualitative examples and evaluation prompts in \cref{suppl:bench_cases,suppl:prompts}.

\section{Dataset Taxonomy}
\label{suppl:pipeline_taxonomy}
To establish our taxonomy, we surveyed prominent datasets, including ImgEdit~\cite{ye2025imgedit}, Step1X-Edit~\cite{liu2025step1x}, and SEED-Data-Edit~\cite{ge2024seeddataedit}, to consolidate and reclassify their editing tasks. We further introduced a novel reasoning category to enhance complex reasoning capabilities and address gaps in existing benchmarks. This process yielded a comprehensive classification of 22 unique editing tasks, spanning from basic object manipulation to complex semantic reasoning, as detailed in \cref{tab:task_taxonomy}.

\input{tables/tab_taxonomy}
\section{Data Pipeline Analysis}
\label{suppl:pipeline_analysis}
\subsection{Comparison with Multi-toolchain Pipelines}
\label{suppl:b1_construction_method}
Automated data curation pipelines~\cite{ye2025imgedit,liu2025step1x,ge2024seeddataedit,zhao2024ultraedit} that rely on a sequence of vision tools are susceptible to error propagation, where inaccuracies from upstream modules are amplified in downstream processes, ultimately degrading dataset quality. To illustrate this, \cref{fig:cases_dataset_sample} presents a qualitative analysis of samples from ImgEdit~\cite{ye2025imgedit} and SEED-Data-Edit~\cite{ge2024seeddataedit}, both of which employ such multi-toolchain architectures. The examples exhibit common artifacts stemming from this approach, including significant blurring and inconsistent color blending at edit boundaries. Furthermore, \cref{fig:cases_dataset_sample}(a) highlights a critical issue of misalignment between the instruction and the resulting edit.

We processed these challenging samples through our proposed pipeline to demonstrate its corrective capabilities. The results show a marked improvement in image quality, with artifacts eliminated and semantic coherence restored. Concurrently, our expert model automatically recaptioned the instructions to align precisely with the actual visual transformations, thereby correcting the instruction-image mismatch and validating the efficacy of our unified verification and refinement process.

\subsection{Comparison of No Edit Filtration Methods}
\label{suppl:b2_no_edit}
A critical step in our post-verification pipeline is the accurate filtration of \textit{No Edit} samples, where the output image is visually identical or nearly identical to the original. Traditional pixel-based metrics like the Structural Similarity Index (SSIM)~\cite{wang2004ssim} are ill-suited for this task, as they lack semantic understanding and are sensitive to imperceptible artifacts introduced by generative models.

\begin{figure*}[!tb]
    \centering
    \includegraphics[width=1\textwidth]{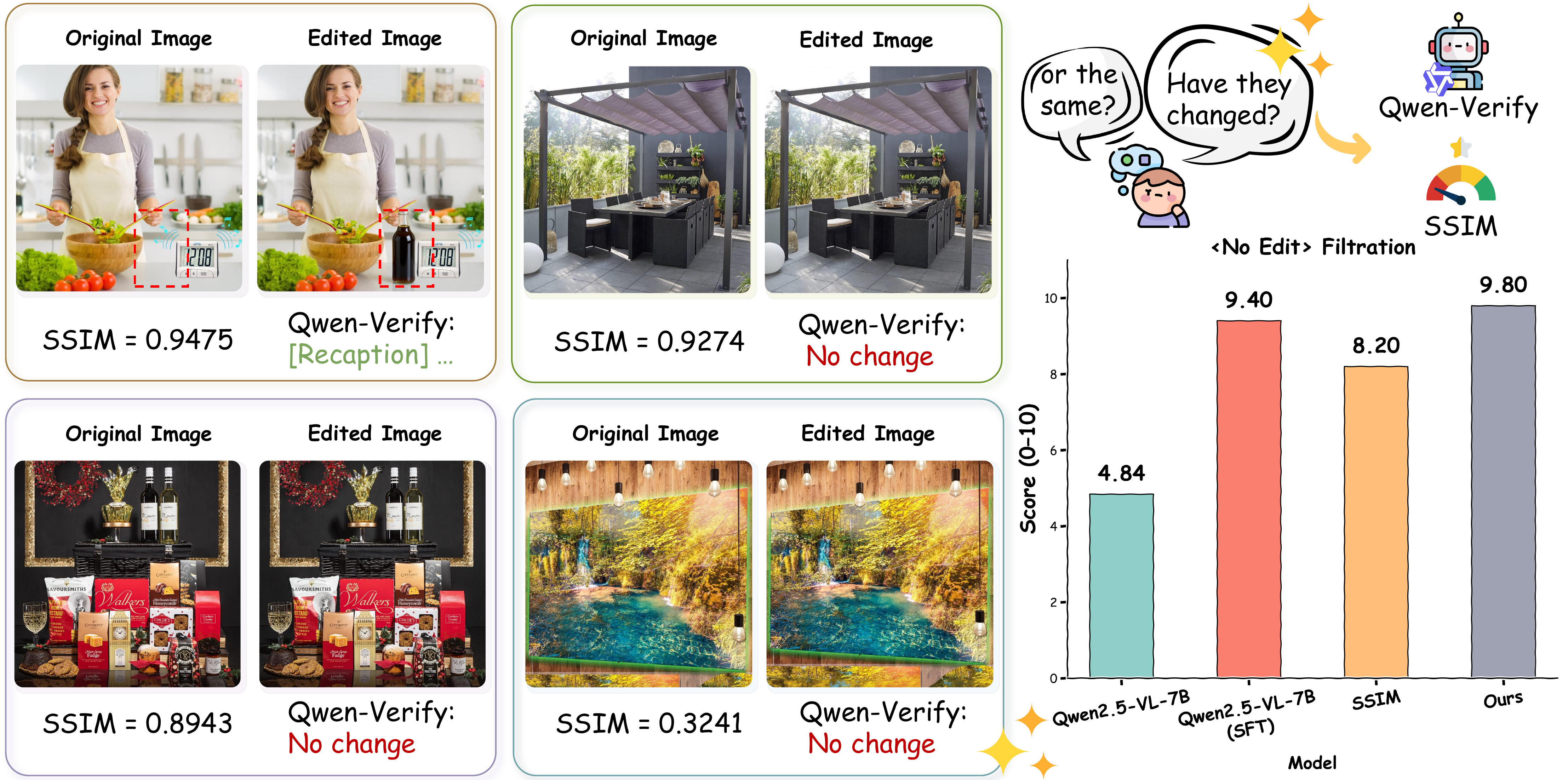}
    % \vspace{-0.5cm}
    \caption{Examples of \textit{No Edit} filtration. The left side shows four examples with detection results from SSIM~\cite{wang2004ssim} and Qwen-Verify. The top-left example has a clear edit (\red{red dashed box}) but receives a high SSIM~\cite{wang2004ssim} score, while the other three visually unchanged pairs receive lower scores. The right side provides a quantitative comparison, confirming that Qwen-Verify performs best at identifying failed edits.}
    % \vspace{-0.6cm}
    \label{fig:cases_filtration}
\end{figure*}

\cref{fig:cases_filtration} illustrates this deficiency. SSIM~\cite{wang2004ssim} proves insensitive to semantically meaningful but visually subtle changes. For instance, the addition of a small bottle (top-left example) yields a high SSIM~\cite{wang2004ssim} of 0.9474, incorrectly suggesting no change occurred. Conversely, SSIM~\cite{wang2004ssim} is overly sensitive to minor, imperceptible pixel shifts inherent to the editing model's process. In the top-right and bottom-left examples, no visible edit was made, yet their SSIM~\cite{wang2004ssim} scores are lower than the sample with a clear object addition, leading to false positives. An even more extreme case (bottom-right example) shows a minor color artifact causing the SSIM~\cite{wang2004ssim} to decline to 0.3241.

In contrast, our expert model demonstrates robust, semantically-aware judgment. It correctly identifies the subtle yet valid edit in the first example while accurately classifying the other visually unchanged pairs as \textit{No Edit}, ignoring trivial pixel-level noise. This capability is crucial for reliable, large-scale data curation. The bar chart in \cref{fig:cases_filtration} provides quantitative validation, confirming that our expert model achieves superior performance in identifying \textit{No Edit} instances compared to baseline methods.

\subsection{Comparison with Other Datasets}
\label{suppl:dataset_comparison}
We provide a comprehensive comparison with existing datasets in \cref{tab:comp_datasets} of the main text. Unlike other datasets, UnicEdit-10M employs a unified post-verification stage that not only filters low-quality samples but also refines instructions to ensure precise alignment with the actual edits. Furthermore, UnicEdit-10M provides extensive data for complex edits, offering broader functional diversity and greater scale than currently available resources.

\section{Expert Model Evaluation}
\label{suppl:expert_eval}
To evaluate Qwen-Verify, we engaged two human experts to curate a total of 390 samples, comprising 300 \textit{Normal} cases, 50 \textit{No Edit} cases, and 40 \textit{Hallucination} cases. For all samples, the experts manually revised any erroneous descriptions to establish reliable ground-truth instructions. To systematically evaluate the performance of Qwen-Verify, we designed an automated evaluation protocol using GPT-4.1. This protocol parses atomic editing tasks from each instruction and computes a score to objectively quantify the alignment between the ground-truth instruction and the instruction generated by our model.

\paragraph{Atomic Task Decomposition.}
We formalize an instruction $P$ as a set of atomic editing tasks, $T_{P}$ . Each atomic task $t_{i}$ is a tuple representing a core operation:
\begin{equation}
    P \xrightarrow{\text{parse}} T_{P} = \{ t_i \}_{i=1}^N, ~ \text{where} ~ t_{i}= (o_{i}, a_{i})
\end{equation}
In this formulation,  $o_{i}$ denotes the target object or region, and $a_{i}$ represents the corresponding edit action.

\paragraph{Alignment Accuracy.}
Given the set of atomic tasks from the ground-truth instruction, $T_{GT}$, and the set from the generated instruction, $T_{GEN}$, we compute an \textit{Alignment Accuracy (Acc.)}. The score measures the coverage of ground-truth tasks while penalizing for hallucinated or redundant tasks. It is defined as:
{\small
\begin{equation}
    \text{Acc}(T_{GEN}, T_{GT}) = \underbrace{\frac{|T_{GT}\cap T_{GEN}|}{|T_{GT}|}}
    _{\text{Coverage (Recall)}}- w \cdot \underbrace{\frac{|T_{GEN}\setminus T_{GT}|}{|T_{GT}|}}
    _{\text{Redundancy Penalty}}
\end{equation}
}
Here, the first term calculates the recall of correctly identified atomic tasks. The second term imposes a penalty, weighted by $w$, for any extraneous tasks generated by the model. In our experiments, we set $w=0.5$. This metric provides a quantitative measure of the precision and fidelity of the generated answers. After validation, the expert model is deployed in our pipeline for large-scale data curation.

\begin{figure*}[!tb]
    \centering
    \includegraphics[width=1\textwidth]{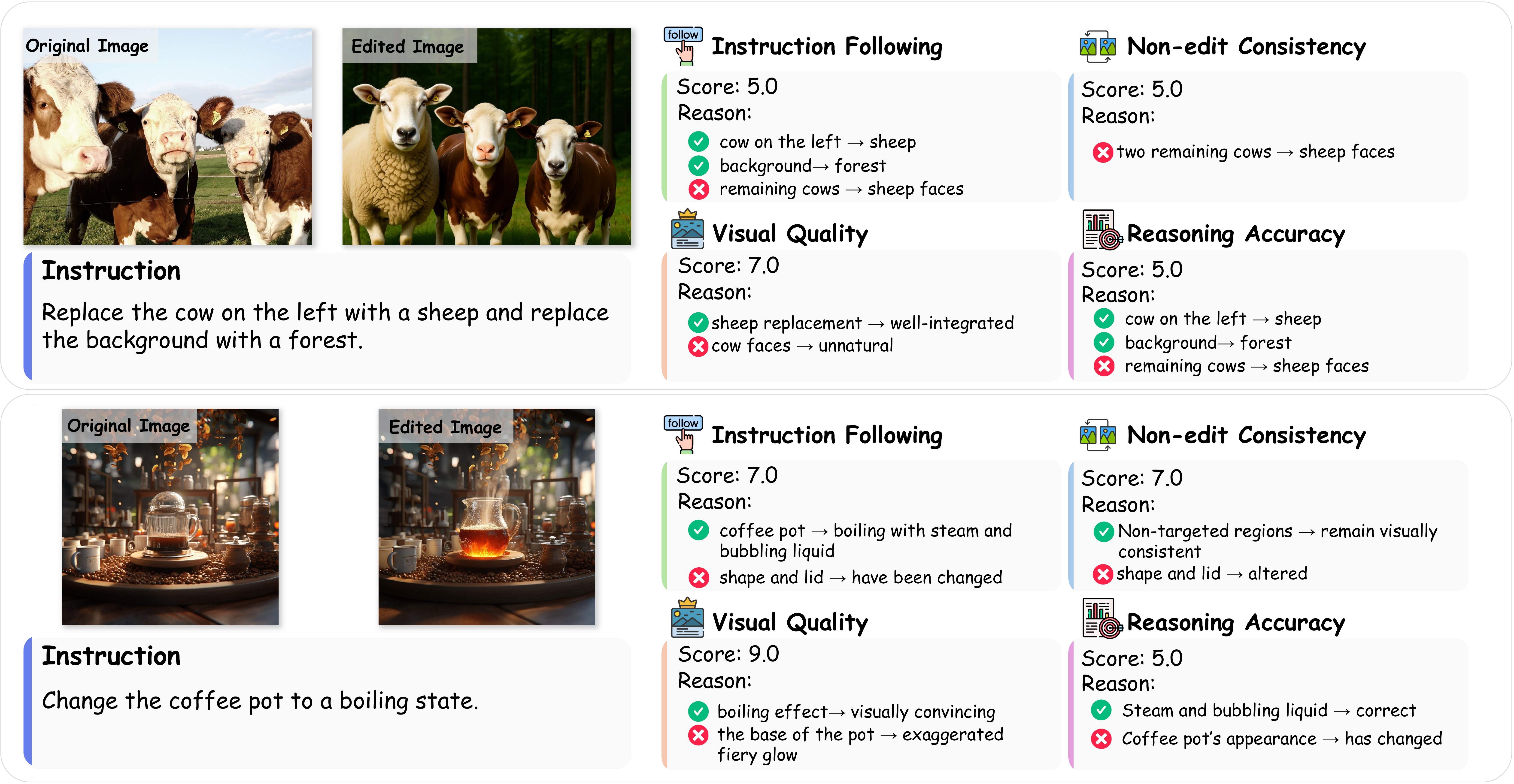}
    % \vspace{-0.5cm}
    \caption{Examples of our evaluation protocol on two complex edit cases. For each example, the edited image pair and instruction are shown on the left. The right side displays the scores for four evaluation dimensions, along with the detailed reasons from the VLM evaluator, which specifies points of credit and deduction.}
    % \vspace{-0.6cm}
    \label{fig:cases_metric_show}
\end{figure*}

\input{tables/tab_benchmarks}

% SC: Semantic Consistency, PQ: Perceptual Quality; IF: Instruction Following, NC: Non-edit Consistency, VQ: Visual Quality.

\section{Validation of Benchmark Metrics}
\label{suppl:analy_metric}
Our evaluation protocol offers a fine-grained assessment by decomposing edit quality into four key dimensions: \textit{Instruction Following (IF)}, \textit{Non-edit Consistency (NC)}, \textit{Visual Quality (VQ)}, and \textit{Reasoning Accuracy (RA)}. This approach offers a more diagnostic alternative to other metrics like the VIEScore~\cite{ku2023viescore}, which can obscure specific failure modes.

\begin{figure}[tbp]
    \centering
    \includegraphics[width=\columnwidth]{
        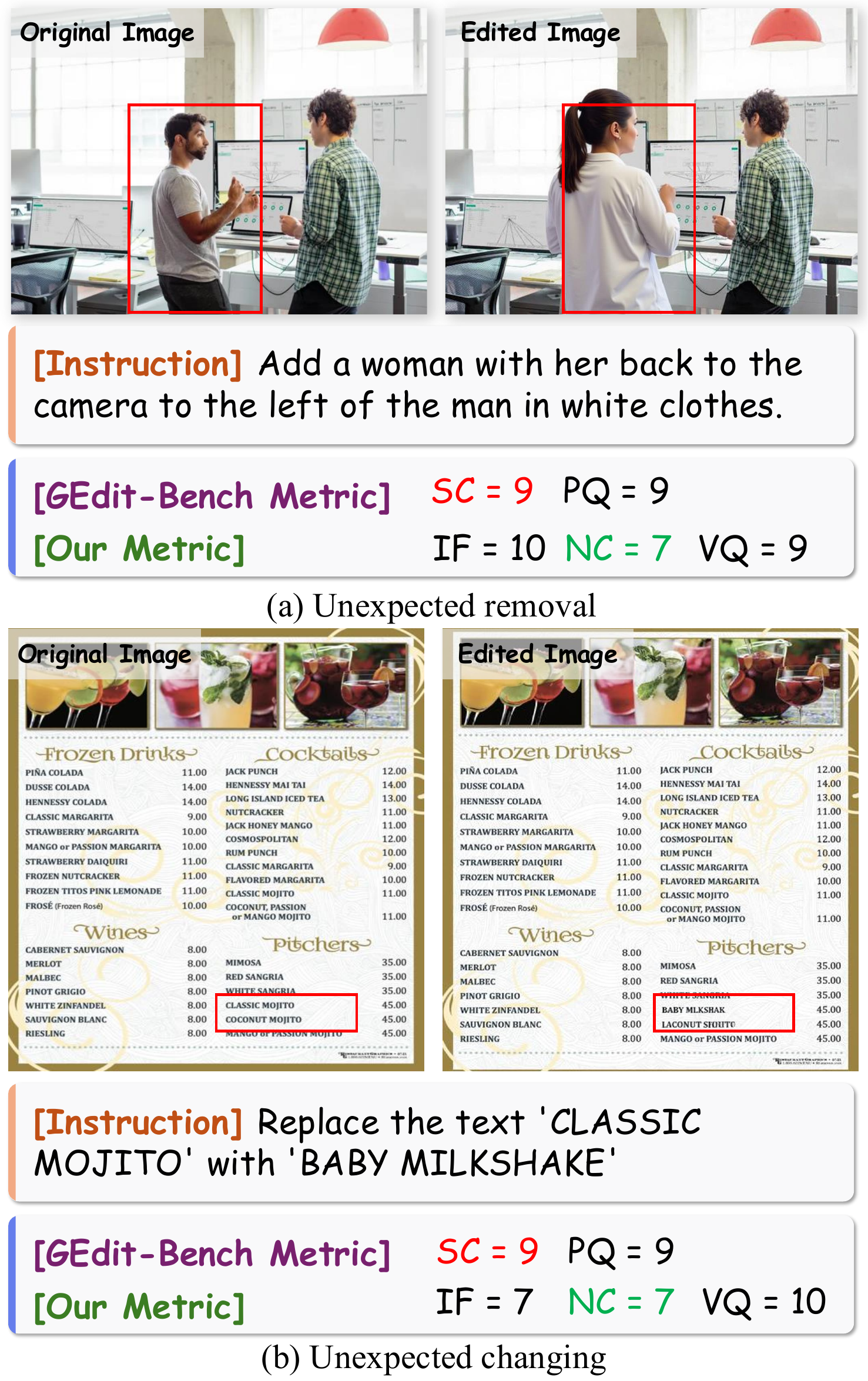
    }
    % \vspace{-0.5cm}
    \caption{Comparison between GEdit-Bench's~\cite{liu2025step1x} metrics and UnicBench's metrics. (a) compares scoring for a case with an unexpected removal. (b) compares scoring for a case with an unexpected text change.}
    % \vspace{-0.6cm}
    \label{fig:cases_metric_comp}
\end{figure}

\paragraph{Comparison with VIEScore.}
As illustrated in \cref{fig:cases_metric_comp}, our protocol demonstrates superior sensitivity to common editing failures. In example (a), the model replaces the man instead of adding a woman to his left. While VIEScore~\cite{ku2023viescore} assigns an SC score of 9, failing to penalize this critical error in non-edit preservation, our NC metric correctly identifies the unintended removal and assigns a lower score of 7. Similarly, in example (b), changing `CLASSIC MOJITO' to `BABY MILKSHAKE' also results in collateral damage to the text below it. VIEScore~\cite{ku2023viescore} again overlooks this whereas our NC metric penalizes the collateral change. These cases show that our fine-grained metrics more reliably assess preservation of non target regions.

\begin{figure}[tbp]
    \centering
    \includegraphics[width=\columnwidth]{figures/cases_facial_v3.pdf}
    % \vspace{-0.5cm}
    \caption{Qualitative comparison of facial consistency. (a) shows examples from GPT-Image-Edit-1.5M~\cite{wang2025gptimageedit}. (b) shows examples from UnicEdit-10M.}
    % \vspace{-0.6cm}
    \label{fig:cases_facial}
\end{figure}

\paragraph{Alignment with Human Judgment.}
To validate that our VLM-based evaluation aligns with human judgment, we analyze its scoring on complex cases in \cref{fig:cases_metric_show}. In the top example, the instruction involves two tasks: changing the leftmost cow to a sheep and the background to a forest. While the model executes these tasks, it incorrectly alters two other cows into sheep. Our protocol accurately diagnoses this partial failure: the IF score is lowered to 5.0 because the instruction was not precisely followed (only the \textit{leftmost} cow was specified), and the NC score is also 5.0, penalizing the unintended modifications to the other animals. The VQ score is 7.0, reflecting good overall image quality but noting the unnatural facial features of the edited animals. Finally, guided by the \textit{reasoning-points list}, the RA metric correctly identifies the erroneous object-type change, resulting in a score of 5.0.

The bottom example tests the model's reasoning capability by requesting to make the coffee pot ``boil". The model correctly adds steam and bubbles but undesirably alters the pot's shape and adds an exaggerated flame. Both IF and NC scores are set to 7.0, acknowledging the partial success while penalizing the unintended shape distortion. The VQ score is reduced due to the unrealistic flame. The RA score is 5.0, recognizing that the concept of ``boiling" was successfully rendered but penalizing the failure to preserve the object's core identity (its shape). These case studies demonstrate that our multi-dimensional metric, as judged by a VLM, provides a comprehensive and human-aligned assessment, effectively pinpointing the specific strengths and weaknesses of an editing model in a single evaluation.

\input{tables/tab_facial}

\section{Analysis of Noisy Data}
\label{suppl:analy_noise}
As outlined in \cref{sec3:dataset_curation}, our initial data generation process produces noisy triplets that fall into three primary categories: (1) \textit{Edit Failure}, where no discernible edit is made; (2) \textit{Instruction-Image Misalignment}, where the visual change deviates from the instruction; and (3) \textit{Others}, which encompasses a range of quality issues.

Our expert model, Qwen-Verify, is specifically trained to address the first two, more prevalent issues by filtering null edits and recaptioning misaligned ones, thereby ensuring semantic and instructional fidelity. For the third category, which includes issues like low aesthetic quality, anatomical deformities (\eg, incorrect limbs on human subjects), and other structural distortions, we employ pre-trained aesthetic scoring models and specialized human-body detectors to automatically flag and remove such low-quality samples. This multi-stage refinement strategy ensures that the final dataset is not only semantically aligned but also meets a high standard of visual quality.

\begin{figure*}[t]
    \centering
    \includegraphics[width=1\textwidth]{
        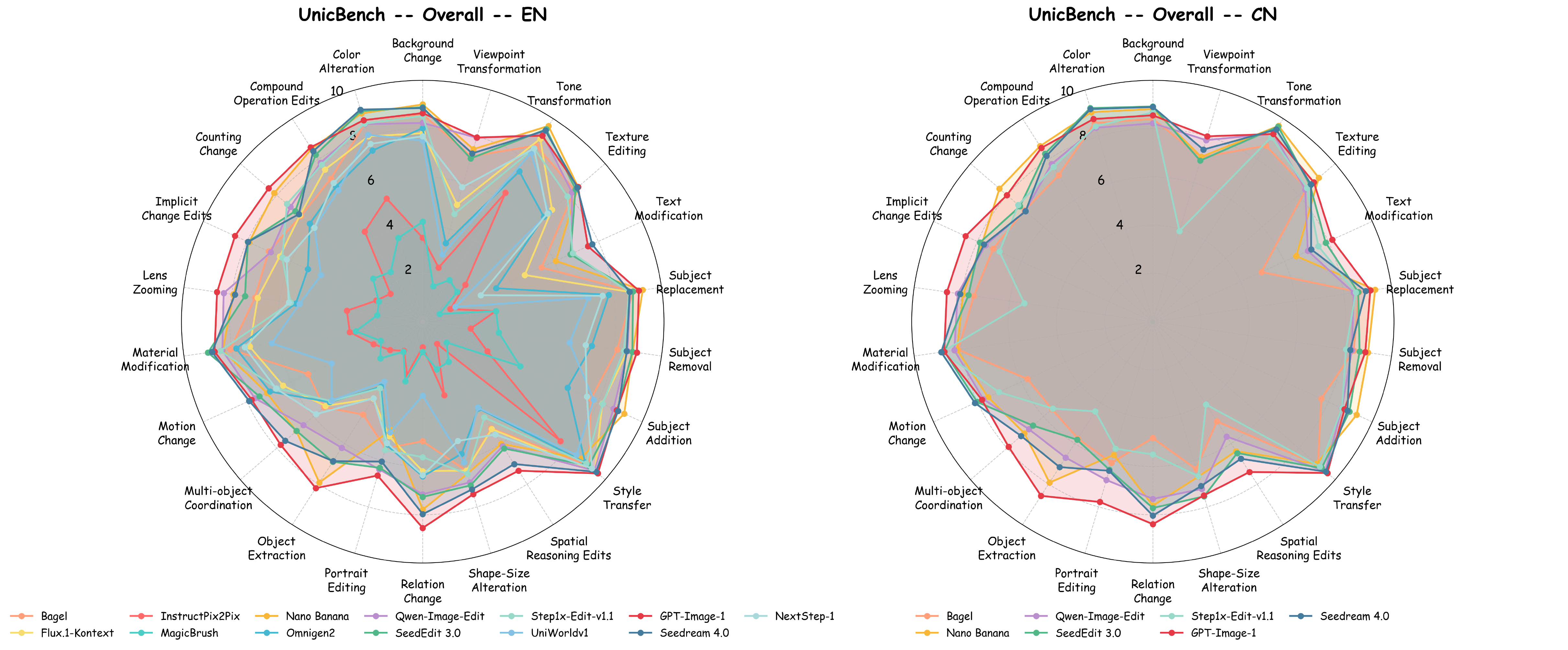
    }
    % \vspace{-0.5cm}
    \caption{Overall score of each model on the sub-tasks in UnicBench, for EN (left) and CN (right) instructions. All results are evaluated by \textit{GPT-4.1}.}
    % \vspace{-0.6cm}
    \label{fig:res_radar_all}
\end{figure*}
\input{tables/tab_benchmark_task_result_en}

\section{Analysis of Facial Consistency}
\label{suppl:facial}
As reported in the main text, our dataset and GPT-Image-Edit-1.5M~\cite{wang2025gptimageedit} achieve comparable Semantic Consistency scores. This prompted a deeper investigation into the qualitative differences between the two datasets. A sampling analysis of GPT-Image-Edit-1.5M~\cite{wang2025gptimageedit} revealed a frequent occurrence of facial identity inconsistency,  a critical issue for practical applications.
To quantify this observation, we conducted a comparison of facial consistency. For portrait image pairs from both datasets, we employed RetinaFace~\cite{deng2020retinaface} for face detection and ArcFace~\cite{deng2019arcface} to extract feature vectors, subsequently calculating the cosine similarity between the faces in the source and edited images. The results, presented in \cref{tab:res_facial}, quantitatively confirm that our dataset maintains a significantly higher face consistency score.
Qualitative examples in \cref{fig:cases_facial} further illustrate this distinction. In the examples from GPT-Image-Edit-1.5M~\cite{wang2025gptimageedit} shown in \cref{fig:cases_facial}~(a), the left image pair shows noticeable changes in the woman’s facial features and face shape, and the right image pair shows a change in the woman’s perceived ethnicity. In contrast, our dataset, as shown in \cref{fig:cases_facial}~(b), exhibits much stronger facial identity preservation. 
This evidence indicates that our data generation pipeline better preserves subject identity throughout the editing process, enhancing the dataset’s overall quality and utility.

\begin{figure*}[!tb]
    \centering
    \includegraphics[width=1\textwidth]{
        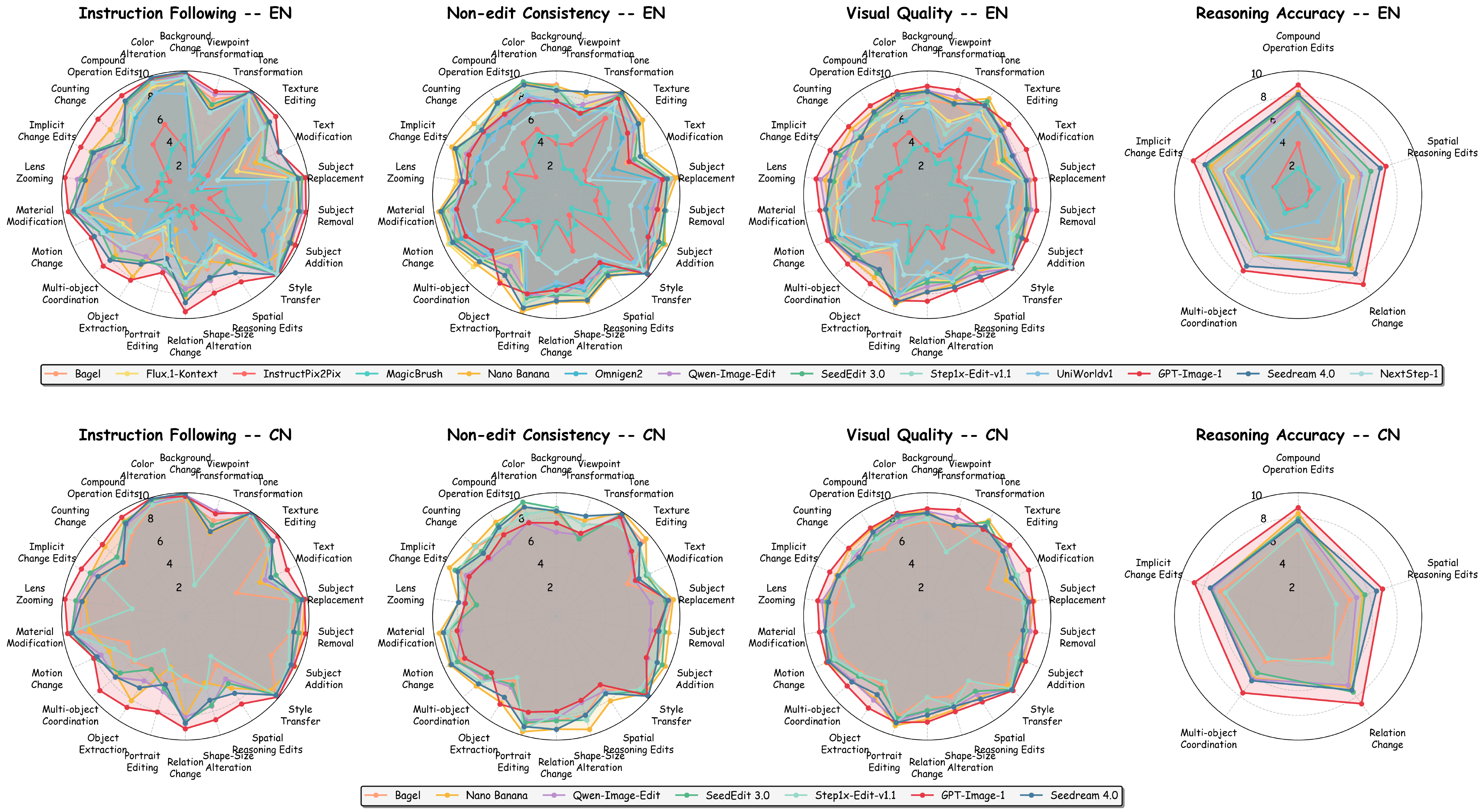
    }
    % \vspace{-0.5cm}
    \caption{Performance of four evaluation dimensions for each sub-task. The top row shows results for EN tasks, and the bottom row shows results for CN tasks. All results are evaluated by \textit{GPT-4.1}.}
    % \vspace{-0.6cm}
    \label{fig:res_radar_each_metric}
\end{figure*}
\input{tables/tab_benchmark_task_result_cn}

\section{Analysis of Benchmark Results}
\label{suppl:bench_res}
\subsection{Overall Analysis}
While \cref{tab:res_bench_overall} of the main text presents the aggregate scores for mainstream models, we visualize their performance breakdown by sub-task in the radar chart in \cref{fig:res_radar_all}. The chart clearly reveals a performance dichotomy: most models perform well on foundational tasks like \textit{Background Change} and \textit{Subject Replacement}, but all models exhibit a significant performance drop on \textit{Viewpoint Transformation}, \textit{Text Modification}, and \textit{Spatial Reasoning Edits}. This indicates that current architectures still lack robust spatial awareness and the ability to precisely edit in-image text.
Furthermore, the visualization highlights a critical gap: open-source models lag behind their closed-source counterparts, particularly on complex reasoning edits. This suggests that the primary bottleneck for open-source models is not basic instruction following, but the advanced inference and world knowledge required for complex manipulation. 

\input{tables/tab_benchmark_task_result_en_qwen}
\input{tables/tab_benchmark_task_result_cn_qwen}

\subsection{Analysis across Different Tasks and Metrics}
A detailed analysis of model performance across different editing tasks is provided in \cref{tab:res_bench_tasks_en} and \cref{tab:res_bench_tasks_cn}. For basic tasks, the performance gap between top open-source models like Qwen-Image-Edit~\cite{wu2025qwenimageedit} and closed-source models is narrow. On some tasks, open-source models even achieve comparable results. For example, Qwen-Image-Edit~\cite{wu2025qwenimageedit} surpasses Nano Banana~\cite{comanici2025nanobanana} in \textit{Attribute Editing} and matches GPT-Image-1~\cite{hurst2024gpt4o} in \textit{Scene Editing}. This suggests that for attribute and scene-level transformations, the capabilities of open-source and closed-source models are converging. However, a substantial performance gap remains in reasoning-intensive tasks, where closed-source models maintain a clear advantage. In this category, Qwen-Image-Edit~\cite{wu2025qwenimageedit}, despite being the SOTA open-source model, lags significantly behind all closed-source models on the IF and RA metrics. This divergence points to the heightened demand for advanced semantic understanding and logical reasoning in complex edits, capabilities that are more densely embedded in closed-source foundation models.

The radar chart in \cref{fig:res_radar_each_metric} visualizes the scores for the four evaluation dimensions and reveals clear performance disparities. On Instruction Following (IF), GPT-Image-1~\cite{hurst2024gpt4o} shows strong semantic understanding and adherence to instructions and it significantly outperforms other models. Notably, nearly all models falter on \textit{Portrait Editing}, \textit{Viewpoint Transformation}, and \textit{Text Modification}, which indicates persistent limitations in portrait editing, spatial understanding, and in-image text manipulation. In contrast, Visual Quality (VQ) scores are more uniform across models, with Qwen-Image-Edit~\cite{wu2025qwenimageedit} performing on par with closed-source models, which suggests that most mainstream models can produce high-quality images. For Non-edit Consistency (NC), GPT-Image-1~\cite{hurst2024gpt4o} lags behind other closed-source models and is surpassed by open-source models such as Qwen-Image-Edit~\cite{wu2025qwenimageedit} and BAGEL~\cite{deng2025bagel}, which implies difficulty in preserving non-target regions during fine-grained edits despite strong instruction understanding. The gaps are most pronounced on Reasoning Accuracy (RA). In EN tasks, closed-source models show clear advantages and open-source models struggle, especially on \textit{Implicit Change Edits} and \textit{Spatial Reasoning Edits}. In CN tasks, the gap narrows, and Qwen-Image-Edit~\cite{wu2025qwenimageedit} matches or exceeds some closed-source models on \textit{Implicit Change Edits} and \textit{Multi-object Coordination}, which suggests that the reasoning abilities of some closed-source models are less robust for Chinese instructions. By integrating basic and complex tasks, UnicBench pinpoints critical limitations in consistency and reasoning. Our findings suggest that future research should focus on aligning semantic understanding with edit execution and on enhancing reasoning capability so that models can leverage world knowledge to satisfy complex user intents.

\subsection{Evaluation on Qwen2.5-VL-72B}
To validate the robustness of our evaluation protocol and mitigate potential biases from a single proprietary evaluator, we employed Qwen2.5-VL-72B~\cite{bai2025qwen2p5vl} as an open-source scoring proxy. The results, detailed in \cref{tab:res_bench_tasks_en_qwen,tab:res_bench_tasks_cn_qwen}, reveal trends that are largely consistent with the GPT-4.1 evaluation, reinforcing the reliability of our findings.
Notably, Qwen-Image-Edit~\cite{wu2025qwenimageedit} maintains its position as the leading open-source model and even surpasses Nano Banana~\cite{comanici2025nanobanana} on EN instructions for all categories except \textit{Reasoning Editing}. This confirms that top-tier open-source models are becoming highly competitive in foundational editing tasks. However, closed-source models retain a decisive advantage in complex edits, particularly for EN instructions. While this gap narrows for CN instructions, closed-source models still maintain a leading position. This underscores that advanced reasoning remains a critical area for the future development of open-source models. The consistent top performance of GPT-Image-1~\cite{hurst2024gpt4o} across both evaluators solidifies its status as the current state-of-the-art image editing model.

\section{Comparison with Other Benchmarks}
\label{suppl:comp_bench}
We compare UnicBench with existing benchmarks in \cref{tab:comp_benchmarks}. As shown, most benchmarks focus primarily on basic edits, offering incomplete coverage of model capabilities. Specialized benchmarks like KRIS-Bench~\cite{wu2025krisbench} target only reasoning tasks and thus cannot serve as general-purpose evaluation tools. While ImgEdit-Bench~\cite{ye2025imgedit} includes complex edits, its coverage is minimal, with only 47 samples across limited categories, making it insufficient for comprehensive measurement. UnicBench addresses these shortcomings by providing balanced coverage of both basic and complex edits across 22 sub-tasks, with all samples manually reviewed for quality. This makes UnicBench a more comprehensive and reliable tool for evaluating the full spectrum of image editing capabilities.

\section{Benchmark Cases}
\label{suppl:bench_cases}
\begin{figure*}[!tb]
    \centering
    \includegraphics[width=1\textwidth]{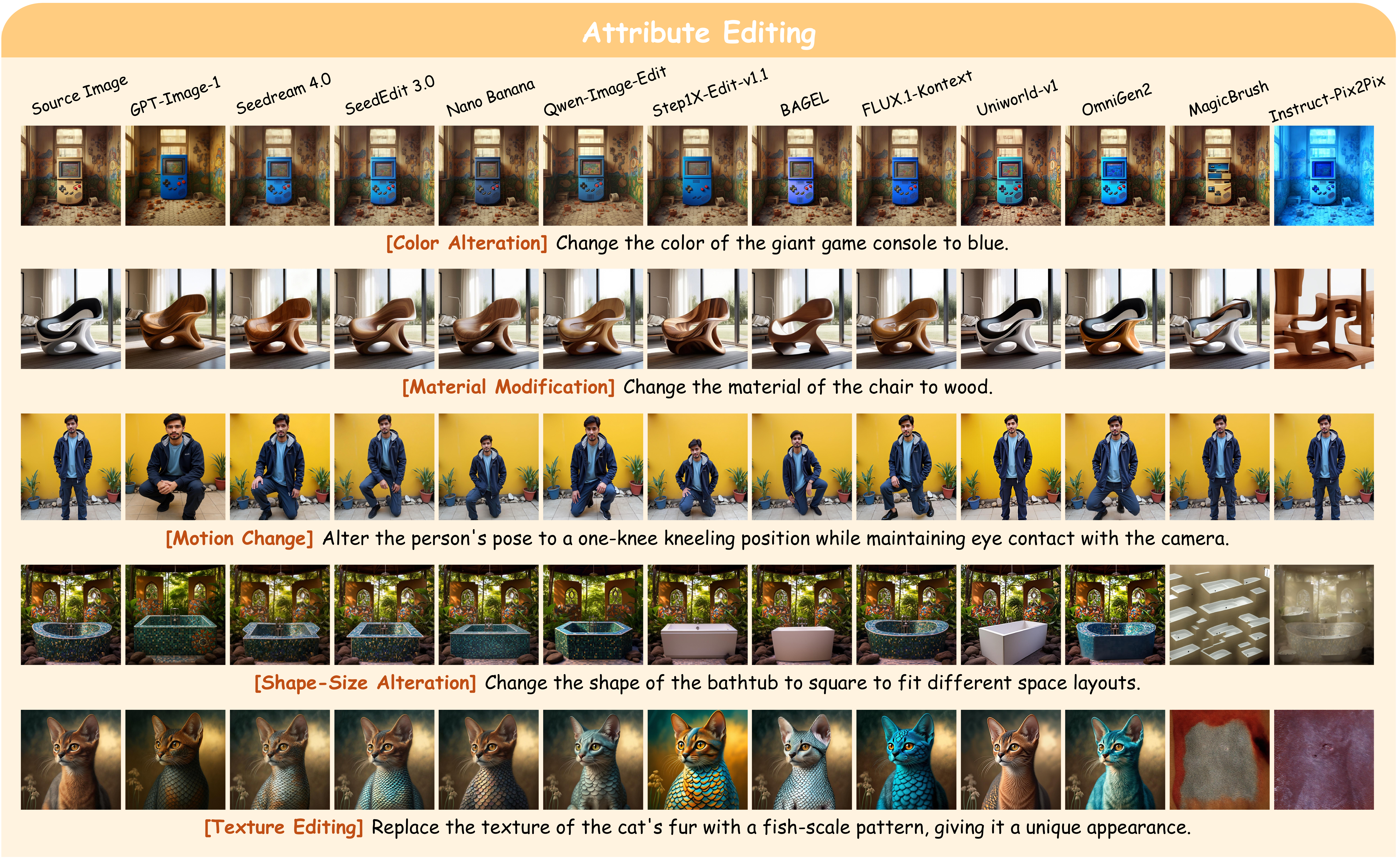}
    % \vspace{-0.5cm}
    \caption{Qualitative results for Attribute Editing tasks on UnicBench (EN).}
    % \vspace{-0.6cm}
    \label{fig:bench_attr_en}
\end{figure*}

\begin{figure*}[!tb]
    \centering
    \includegraphics[width=1\textwidth]{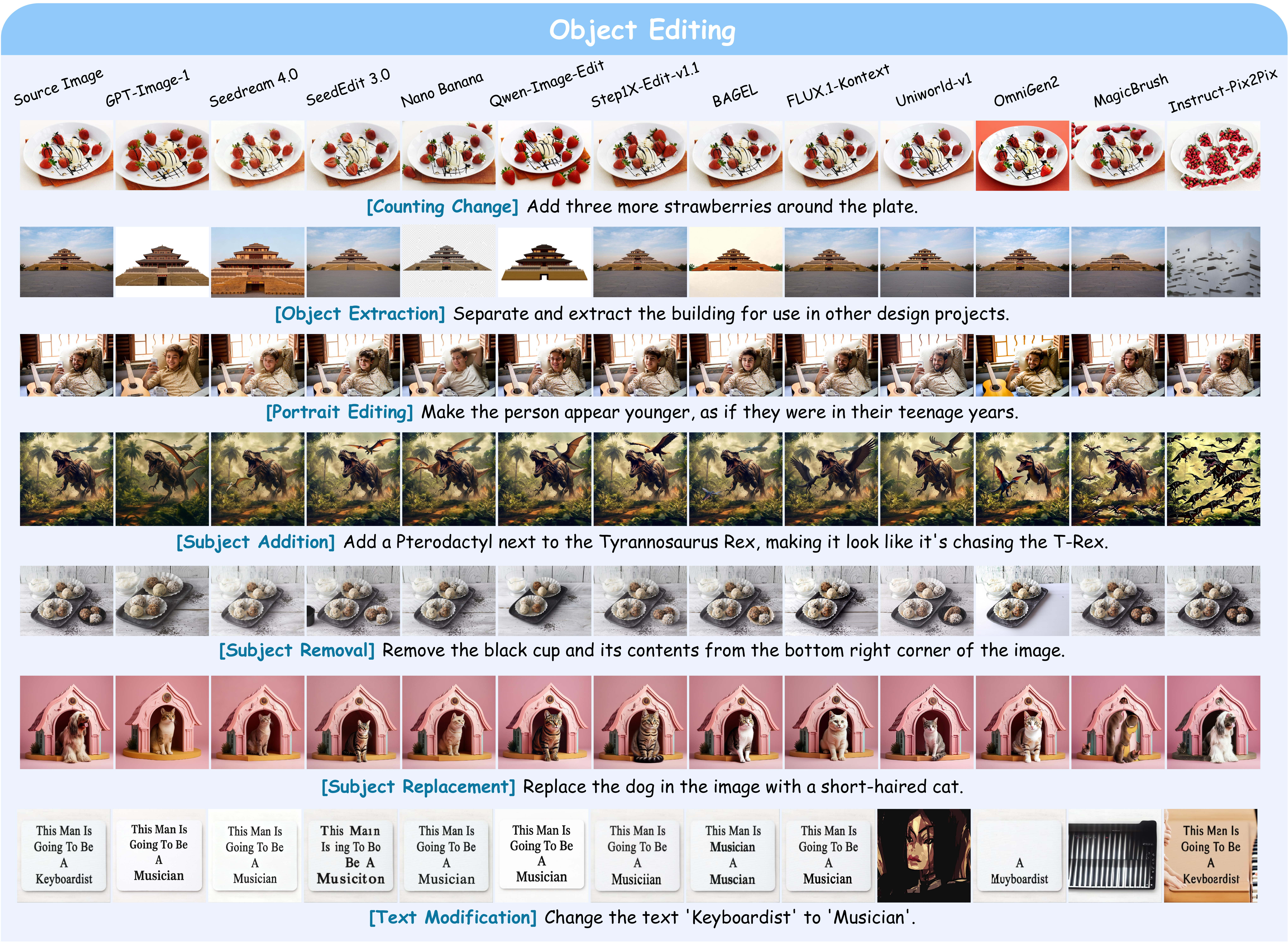}
    % \vspace{-0.5cm}
    \caption{Qualitative results for Object Editing tasks on UnicBench (EN).}
    % \vspace{-0.6cm}
    \label{fig:bench_obje_en}
\end{figure*}

\begin{figure*}[!tb]
    \centering
    \includegraphics[width=1\textwidth]{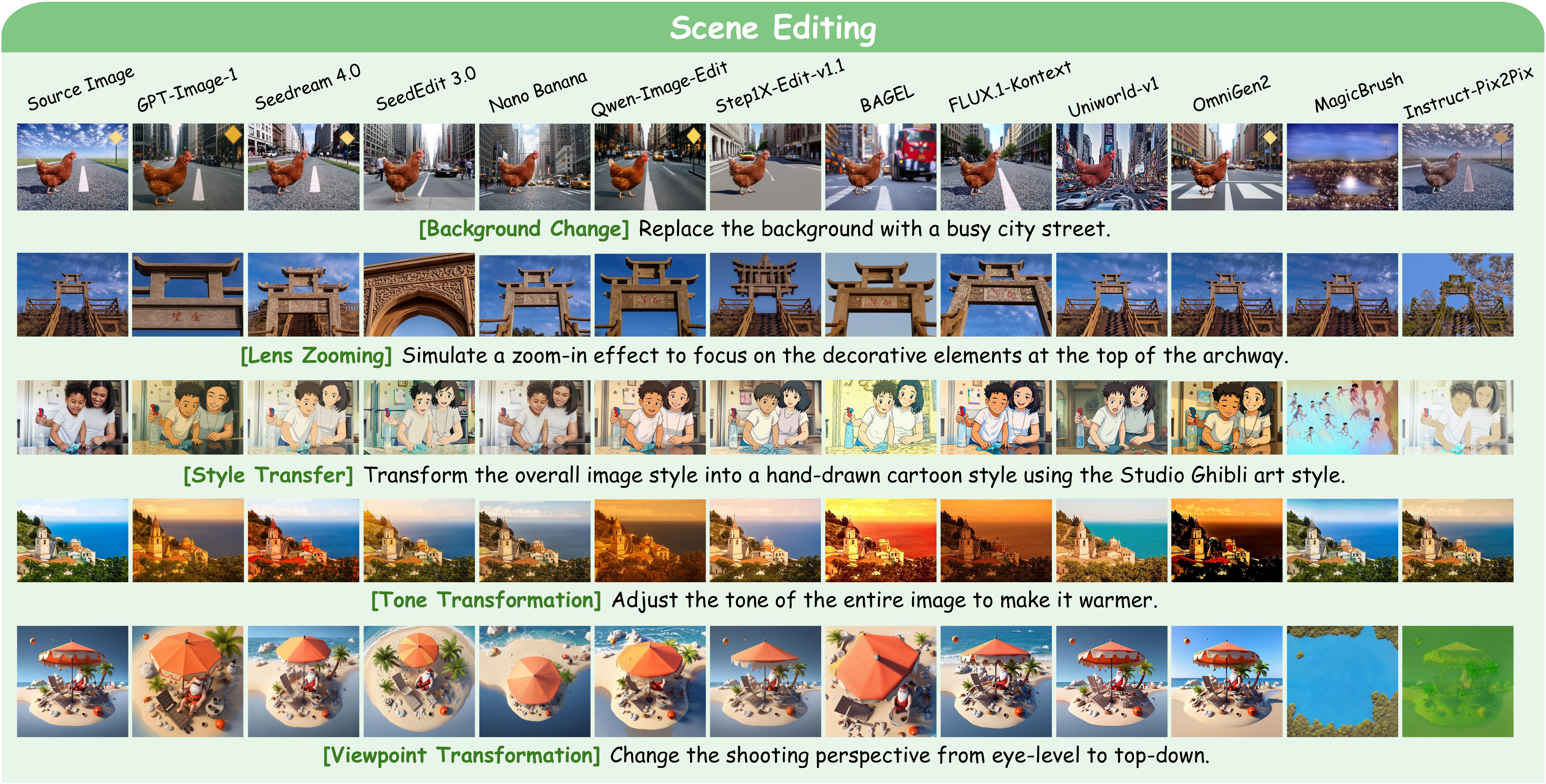}
    % \vspace{-0.5cm}
    \caption{Qualitative results for Scene Editing tasks on UnicBench (EN).}
    % \vspace{-0.6cm}
    \label{fig:bench_scene_en}
\end{figure*}

\begin{figure*}[!tb]
    \centering
    \includegraphics[width=1\textwidth]{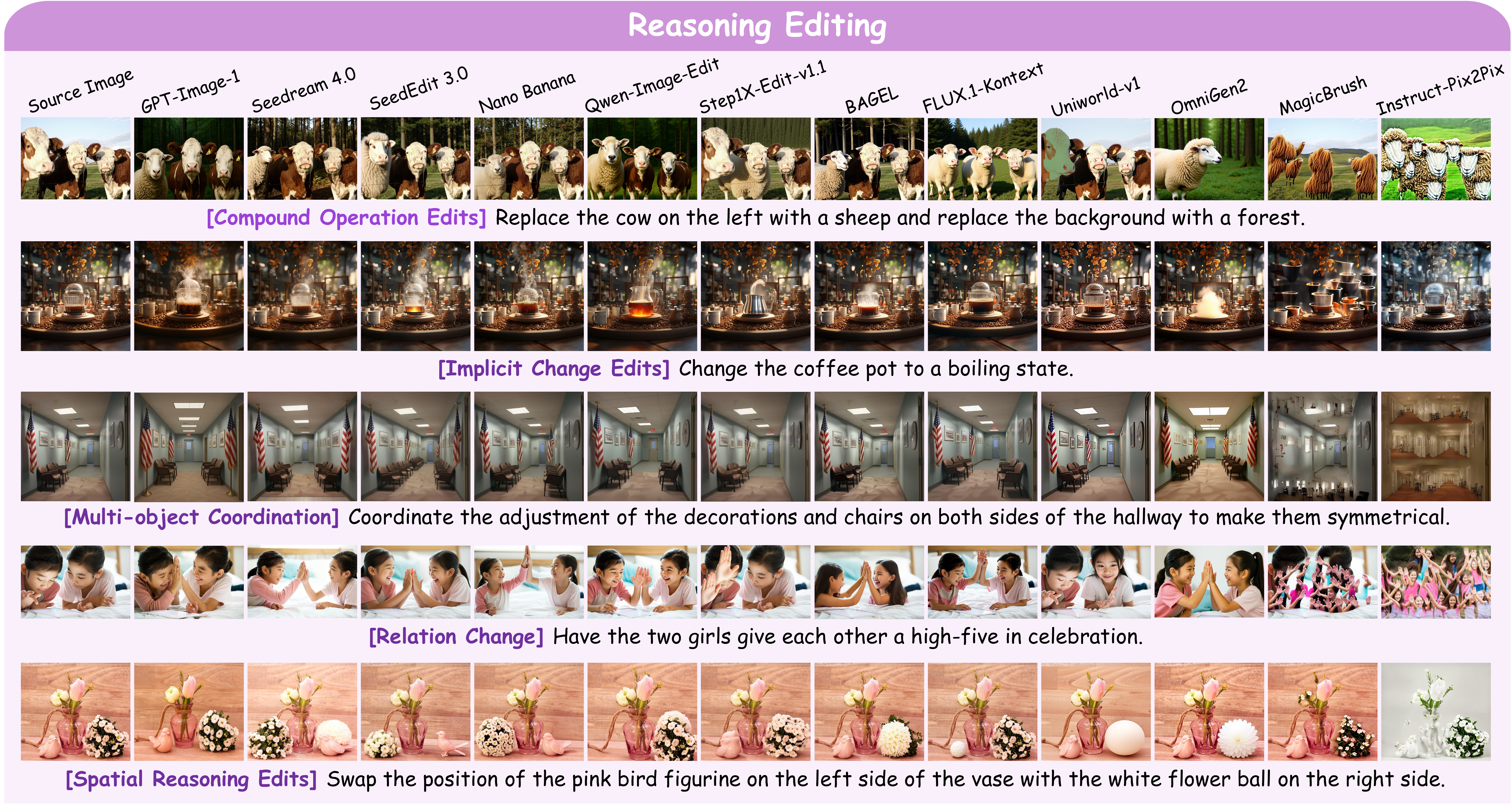}
    % \vspace{-0.5cm}
    \caption{Qualitative results for Reasoning Editing tasks on UnicBench (EN).}
    % \vspace{-0.6cm}
    \label{fig:bench_reason_en}
\end{figure*}

In this section, we present additional qualitative results. We showcase examples from mainstream models across our four main editing categories: \textit{Attribute Editing} (\cref{fig:bench_attr_en}), \textit{Object Editing} (\cref{fig:bench_obje_en}), \textit{Scene Editing} (\cref{fig:bench_scene_en}), and \textit{Reasoning Editing} (\cref{fig:bench_reason_en}). In each figure, the first column displays the original image, and the subsequent columns show the outputs from the evaluated models.
% In this section, we present extensive qualitative comparisons to further substantiate our quantitative findings. We showcase representative outputs from 12 models across our four editing categories: \textit{Attribute Editing}, \textit{Object Editing}, \textit{Scene Editing}, and \textit{Reasoning Editing}. To demonstrate the multilingual robustness of the evaluated models, we provide examples for both English (EN) and Chinese (CN) instructions. Specifically, \cref{fig:bench_attr_en,fig:bench_obje_en,fig:bench_scene_en,fig:bench_reason_en} illustrate the results for EN tasks, while \cref{fig:bench_attr_cn,fig:bench_obje_cn,fig:bench_scene_cn,fig:bench_reason_cn} display the corresponding results for CN tasks. In each figure, the leftmost column presents the original image, followed by the editing results generated by each model.

\section{Evaluation Prompts}
\label{suppl:prompts}
\cref{prompt:if,prompt:nc,prompt:vq,prompt:ra} detail the instructions provided to the VLM evaluator for assessing \textit{Instruction Following}, \textit{Non-edit Consistency}, \textit{Visual Quality}, and \textit{Reasoning Accuracy}, respectively. For each dimension, the evaluation is based on a set of inputs including the original image, the edited image, and the edit instruction. To facilitate a more accurate assessment of complex tasks, the evaluation of \textit{Reasoning Accuracy} is uniquely supplemented with a list of ``Reasoning Points" that guide the VLM in its evaluation process.

\onecolumn
\input{prompts/prom_if}

\input{prompts/prom_nc}
\input{prompts/prom_vq}
\input{prompts/prom_ra}
\twocolumn

%% file: tables/tab_taxonomy.tex
\begin{table*}[ht]
\centering
\caption{Definitions of editing taxonomy of UnicEdit-10M.}
\label{tab:task_taxonomy}
% resizebox to ensure the table fits the page width
\resizebox{\textwidth}{!}{
\begin{tabular}{lll}
\toprule[2pt] % Top thick rule (2pt)
\textbf{Task} & \textbf{Subtask} & \textbf{Description} \\
\midrule\midrule % Double rule under the header, as per your reference style
\multirow{7}{*}{Object Editing} & Subject Addition & Adding a new object or person to the image. \\
& Subject Removal & Removing a specified object or person from the image. \\
& Subject Replacement & Replacing an object with another. \\
& Object Extraction & Isolating and extracting a target object from its background. \\
& Counting Change & Modifying the number of objects in the image. \\
& Text Modification & Editing textual content within the image. \\
& Portrait Editing & Enhancing or modifying facial features. \\
\midrule % Rule to separate the groups
\multirow{5}{*}{Attribute Editing} & Color Alteration & Changing the color of an object or region. \\
& Material Modification & Changing the material properties of an object. \\
& Motion Change & Adjusting the dynamic pose or action of an object. \\
& Texture Editing & Modifying the surface texture details of an object. \\
& Shape-Size Alteration & Changing the shape or size of an object. \\
\midrule % Rule to separate the groups
\multirow{5}{*}{Scene Editing} & Background Change & Replacing or modifying the image background. \\
& Style Transfer & Converting the overall image to a target artistic style. \\
& Tone Transformation & Adjusting the image's color tone and atmosphere. \\
& Viewpoint Transformation & Changing the camera's viewpoint or position. \\
& Lens Zooming & Simulating an optical lens zoom effect. \\
\midrule % Rule to separate the groups
\multirow{5}{*}{Reasoning Editing} & Spatial Reasoning & Moving objects based on spatial logic. \\
& Multi-Object Coordination & Coordinately modifying the attributes or positions of multiple objects. \\
& Compound Operations & Executing multiple, combined editing operations. \\
& Relation Change & Modifying the interactive relationship between objects. \\
& Implicit Change Edits & Inferring the actual editing task based on context and real-world knowledge. \\
\bottomrule[2pt] % Bottom thick rule (2pt)
\end{tabular}
}
\end{table*}

%% file: tables/tab_benchmarks.tex
\begin{table*}[t]
\centering
\caption{Comparison of different image editing benchmarks.}
\label{tab:comp_benchmarks}
% Use resizebox to ensure the table fits within the text width
\setlength{\tabcolsep}{14pt}
\resizebox{\textwidth}{!}{
\begin{tabular}{l|ccccccc}
\toprule[2pt] % Top thick rule (2pt)
\textbf{Benchmarks} & \textbf{\#Size} & \textbf{\#Subtasks} & \textbf{Human Filtering} & \textbf{Basic Edit} & \textbf{Complex Edit} & \textbf{Public Available} \\
\midrule\midrule % Double rule under the header, as per your reference style
EditBench~\cite{wang2023editbench}   & 240   & 1  & \xmark & \cmark & \xmark & \cmark \\
EmuEdit~\cite{sheynin2024emuedit}    & 3055  & 7  & \xmark & \cmark & \xmark & \cmark \\
HIVE~\cite{zhang2024hive}            & 1000  & 1  & \cmark & \cmark & \xmark & \cmark \\
HQ-Edit~\cite{hui2024hqedit}         & 1640  & 7  & \xmark & \cmark & \xmark & \cmark \\
MagicBrush~\cite{zhang2023magicbrush}& 1053  & 7  & \cmark & \cmark & \xmark & \cmark \\
AnyEdit-Test~\cite{yu2025anyedit}    & 1250  & 25 & \xmark & \cmark & \xmark & \cmark \\
ICE-Bench~\cite{pan2025icebench}     & 6538  & 31 & \cmark & \cmark & \xmark & \xmark \\
ImgEdit-Bench~\cite{ye2025imgedit}   & 737   & 9  & \cmark & \cmark & \cmark & \cmark \\
GEdit-Bench~\cite{liu2025step1x}     & 606   & 11 & \cmark & \cmark & \xmark & \cmark \\
KRIS-Bench~\cite{wu2025krisbench}    & 1267  & 22 & \cmark & \xmark & \cmark & \cmark \\
\midrule % Single rule before the "Ours" row, as per your reference style
\textbf{UnicBench}                       & 1100  & 22 & \cmark & \cmark & \cmark & \cmark \\
\bottomrule[2pt] % Bottom thick rule (2pt)
\end{tabular}
}
\end{table*}

%% file: tables/tab_facial.tex
\begin{table}[b]
\centering
\caption{Comparison of facial consistency between GPT-Image-Edit-1.5M~\cite{wang2025gptimageedit} and UnicEdit-10M.}
\label{tab:res_facial}
% Using resizebox for style consistency, width is adjusted for a smaller table
\resizebox{0.4\textwidth}{!}{
\begin{tabular}{l|c}
\toprule[2pt] % Top thick rule (2pt)
\textbf{Datasets} & \textbf{Facial Consistency} \\
\midrule\midrule % Double rule under the header, as per your reference style
GPT-Image-Edit-1.5M~\cite{wang2025gptimageedit} & 0.3025 \\
\textbf{Ours} & \textbf{0.8911} \\
\bottomrule[2pt] % Bottom thick rule (2pt)
\end{tabular}
}
\end{table}

%% file: tables/tab_benchmark_task_result_en.tex
\begin{table*}[ht]
\centering
\caption{Detailed performance across different editing tasks (EN). The performance of open-source and closed-source models is separately marked with the best performance in \textbf{bold}, and the second best \underline{underlined}. All results are evaluated by \textit{GPT-4.1}.}
\label{tab:res_bench_tasks_en}
% resizebox is essential here to make the very wide table fit the page width
\resizebox{\textwidth}{!}{
\begin{tabular}{l|cccc|cccc|cccc|ccccc}
\toprule[2pt] % Top thick rule (2pt)
\multirow{2}{*}{\textbf{Model}} & \multicolumn{4}{c|}{\textbf{Attribute Editing}} & \multicolumn{4}{c|}{\textbf{Object Editing}} & \multicolumn{4}{c|}{\textbf{Scene Editing}} & \multicolumn{5}{c}{\textbf{Reasoning Editing}} \\
\cmidrule(lr){2-5} \cmidrule(lr){6-9} \cmidrule(lr){10-13} \cmidrule(lr){14-18} % Rules under the top-level headers
& \textbf{IF} & \textbf{NC} & \textbf{VQ} & \textbf{Overall} & \textbf{IF} & \textbf{NC} & \textbf{VQ} & \textbf{Overall} & \textbf{IF} & \textbf{NC} & \textbf{VQ} & \textbf{Overall} & \textbf{IF} & \textbf{NC} & \textbf{VQ} & \textbf{RA} & \textbf{Overall} \\
\midrule\midrule % Double rule under the header, as per your reference style
\rowcolor{gray!20}
\multicolumn{18}{c}{\textit{Open-Source Models}} \\ % Sub-header for the group
Instruct-Pix2Pix~\cite{brooks2023instructpix2pix}    & 3.1320 & 4.3880 & 4.3840 & 3.2273 & 2.0143 & 3.3152 & 3.3754 & 2.1075 & 4.2880 & 5.7240 & 4.8600 & 4.5773 & 2.3080 & 3.2760 & 3.4840 & 1.9560 & 2.0990 \\
MagicBrush~\cite{zhang2023magicbrush}                & 2.3655 & 3.6466 & 3.6185 & 2.4559 & 2.5229 & 3.7000 & 3.6514 & 2.6074 & 2.3120 & 2.7320 & 3.1960 & 2.1907 & 2.0880 & 3.3360 & 3.2800 & 1.7240 & 2.0027 \\
OmniGen2~\cite{wu2025omnigen2}                       & 7.1600 & 7.9080 & 6.8640 & 6.9011 & 5.5514 & 7.1400 & 6.3514 & 5.6477 & 6.6880 & 7.9680 & 6.4560 & 6.6075 & 5.8600 & 7.1160 & 6.3400 & 5.1240 & 5.5329 \\
UniWorld-v1~\cite{lin2025uniworldv1}                 & 5.8080 & 8.1480 & 7.0200 & 6.3074 & 4.8457 & 6.6886 & 6.1486 & 5.1441 & 6.1360 & 7.5480 & 6.8840 & 6.4598 & 4.6160 & 7.1000 & 6.0120 & 4.0160 & 4.6767 \\
FLUX.1-Kontext~\cite{batifol2025flux}                & 6.8760 & \textbf{8.9160} & 7.4880 & 7.0269 & 6.4286 & \textbf{8.1486} & 7.2800 & 6.5028 & 7.7040 & \textbf{8.4960} & {\ul 7.6200} & 7.6453 & 6.2320 & \textbf{8.4560} & {\ul 7.0840} & 5.5040 & {\ul 6.1634} \\
NextStep-1~\cite{nextstepteam2025nextstep1}          & {\ul 7.8600} & 6.3160 & 6.9320 & 6.7591 & 6.3000 & 5.5229 & 6.0029 & 5.6326 & 8.4056 & 7.1165 & 7.0241 & 7.3901 & {\ul 6.7360} & 6.0400 & 6.2480 & {\ul 6.2920} & 6.1626 \\
BAGEL~\cite{deng2025bagel}                           & 7.5440 & {\ul 8.8120} & 7.3400 & 7.3184 & 6.6343 & 7.9771 & 7.0800 & 6.6784 & {\ul 8.9200} & 8.2400 & 7.5960 & {\ul 8.1083} & 6.1440 & 7.8520 & 6.5640 & 5.2600 & 5.9326 \\
Step1X-Edit-v1.1~\cite{liu2025step1x}                & 7.7560 & 8.7280 & {\ul 7.8440} & {\ul 7.6342} & {\ul 6.7886} & {\ul 8.1400} & {\ul 7.4314} & {\ul 6.9090} & 7.4880 & 7.9280 & 7.2280 & 7.2437 & 6.0280 & {\ul 8.0480} & 6.8120 & 5.0400 & 5.8982 \\
Qwen-Image-Edit~\cite{wu2025qwenimageedit}           & \textbf{8.3960} & 8.2800 & \textbf{8.0720} & \textbf{7.9378} & \textbf{7.8400} & 7.8029 & \textbf{8.0657} & \textbf{7.5128} & \textbf{9.2960} & {\ul 8.3440} & \textbf{8.5800} & \textbf{8.6277} & \textbf{7.4360} & 7.7680 & \textbf{7.5840} & \textbf{6.4480} & \textbf{6.9168} \\
\midrule % Rule to separate the two groups
\rowcolor{gray!20}
\multicolumn{18}{c}{\textit{Closed-source Models}} \\ % Sub-header for the group
Nano Banana~\cite{comanici2025nanobanana}            & 7.7000 & \textbf{9.3040} & {\ul 8.2160} & 7.9117 & 7.6319 & \textbf{9.0348} & {\ul 8.2174} & {\ul 7.7034} & 8.7400 & \textbf{8.8760} & {\ul 8.4800} & 8.5360 & 7.9600 & \textbf{8.6880} & 7.8600 & 6.8680 & 7.4324 \\
SeedEdit 3.0~\cite{wang2025seededit3}                & 8.5265 & 8.9959 & 8.0082 & 8.2087 & 7.7609 & 8.3382 & 7.7230 & 7.6277 & {\ul 9.1215} & 8.1417 & 8.2672 & 8.4011 & 7.8785 & 8.2632 & 7.4049 & 6.9393 & 7.3269 \\
Seedream 4.0~\cite{seedream2025seedream4p0}          & {\ul 8.6160} & {\ul 9.1040} & 8.1880 & \textbf{8.3225} & {\ul 7.8257} & {\ul 8.5286} & 7.9200 & 7.6482 & 8.9960 & {\ul 8.8560} & 8.3280 & {\ul 8.5813} & {\ul 8.2880} & {\ul 8.4680} & {\ul 7.9200} & {\ul 7.5960} & {\ul 7.7770} \\
GPT-Image-1~\cite{hurst2024gpt4o}                    & \textbf{9.1532} & 7.7957 & \textbf{8.6000} & {\ul 8.2475} & \textbf{8.8448} & 8.0090 & \textbf{8.7373} & \textbf{8.2422} & \textbf{9.6680} & 8.1618 & \textbf{8.8631} & \textbf{8.8009} & \textbf{9.0705} & 7.3172 & \textbf{8.4978} & \textbf{8.3392} & \textbf{8.1576} \\
\bottomrule[2pt] % Bottom thick rule (2pt)
\end{tabular}
}
\end{table*}

%% file: tables/tab_benchmark_task_result_cn.tex
\begin{table*}[ht]
\centering
\caption{Detailed performance across different editing tasks (CN). The performance of open-source and closed-source models is separately marked with the best performance in \textbf{bold}, and the second best \underline{underlined}. All results are evaluated by \textit{GPT-4.1}.}
\label{tab:res_bench_tasks_cn}
% resizebox is essential here to make the very wide table fit the page width
\resizebox{\textwidth}{!}{
\begin{tabular}{l|cccc|cccc|cccc|ccccc}
\toprule[2pt] % Top thick rule (2pt)
\multirow{2}{*}{\textbf{Model}} & \multicolumn{4}{c|}{\textbf{Attribute Editing}} & \multicolumn{4}{c|}{\textbf{Object Editing}} & \multicolumn{4}{c|}{\textbf{Scene Editing}} & \multicolumn{5}{c}{\textbf{Reasoning Editing}} \\
\cmidrule(lr){2-5} \cmidrule(lr){6-9} \cmidrule(lr){10-13} \cmidrule(lr){14-18} % Rules under the top-level headers
& \textbf{IF} & \textbf{NC} & \textbf{VQ} & \textbf{Overall} & \textbf{IF} & \textbf{NC} & \textbf{VQ} & \textbf{Overall} & \textbf{IF} & \textbf{NC} & \textbf{VQ} & \textbf{Overall} & \textbf{IF} & \textbf{NC} & \textbf{VQ} & \textbf{RA} & \textbf{Overall} \\
\midrule\midrule % Double rule under the header, as per your reference style
\rowcolor{gray!20}
\multicolumn{18}{c}{\textit{Open-source Models}} \\ % Sub-header for the group
BAGEL~\cite{deng2025bagel}                   & 7.5560 & {\ul 8.8120} & 7.5080 & 7.4498 & 6.7114 & {\ul 8.1829} & 7.3029 & 6.9385 & {\ul 9.0360} & {\ul 8.2000} & {\ul 7.6880} & {\ul 8.1469} & {\ul 6.1400} & {\ul 7.9840} & 6.7520 & {\ul 5.2840} & {\ul 5.9540} \\
Step1X-Edit-v1.1~\cite{liu2025step1x}        & {\ul 8.0080} & \textbf{8.8640} & {\ul 7.9360} & {\ul 7.8829} & {\ul 6.9229} & \textbf{8.4171} & {\ul 7.8429} & {\ul 7.1484} & 7.2600 & \textbf{8.2800} & 7.4000 & 7.2448 & 5.9640 & \textbf{8.0840} & {\ul 6.9480} & 5.0560 & 5.9374 \\
Qwen-Image-Edit~\cite{wu2025qwenimageedit}   & \textbf{8.6200} & 7.9400 & \textbf{8.2240} & \textbf{8.0001} & \textbf{7.8971} & 7.5971 & \textbf{8.2286} & \textbf{7.5682} & \textbf{9.4200} & 8.1120 & \textbf{8.5920} & \textbf{8.5920} & \textbf{7.7400} & 7.6320 & \textbf{7.7960} & \textbf{6.6560} & \textbf{7.0400} \\
\midrule % Rule to separate the two groups
\rowcolor{gray!20}
\multicolumn{18}{c}{\textit{Closed-source Models}} \\ % Sub-header for the group
Nano Banana~\cite{comanici2025nanobanana}    & 7.8040 & \textbf{9.4760} & {\ul 8.4640} & 8.1122 & {\ul 8.0836} & \textbf{9.0317} & {\ul 8.3026} & {\ul 8.0299} & 8.7800 & {\ul 8.7960} & 8.4480 & 8.5037 & 7.9800 & \textbf{8.8760} & {\ul 8.1120} & 6.8960 & 7.4994 \\
SeedEdit 3.0~\cite{wang2025seededit3}        & {\ul 8.7854} & 9.0040 & 8.1457 & \textbf{8.4379} & 7.8006 & {\ul 8.3900} & 7.8240 & 7.6794 & {\ul 9.2418} & 8.2213 & {\ul 8.4590} & 8.5126 & 7.8807 & 8.2016 & 7.5473 & 6.8395 & 7.3807 \\
Seedream 4.0~\cite{seedream2025seedream4p0}  & 8.6840 & {\ul 9.0080} & 8.2760 & {\ul 8.3881} & 7.8114 & 8.3771 & 7.9657 & 7.6935 & 9.0520 & \textbf{8.9280} & 8.3600 & {\ul 8.6731} & {\ul 8.0320} & {\ul 8.4400} & 8.0120 & {\ul 7.1240} & {\ul 7.5765} \\
GPT-Image-1~\cite{hurst2024gpt4o}            & \textbf{9.2876} & 7.8541 & \textbf{8.6481} & 8.3420 & \textbf{9.1381} & 8.0150 & \textbf{8.7688} & \textbf{8.4681} & \textbf{9.6292} & 8.2625 & \textbf{8.8375} & \textbf{8.8071} & \textbf{9.0925} & 7.3524 & \textbf{8.4978} & \textbf{8.2247} & \textbf{8.1594} \\
\bottomrule[2pt] % Bottom thick rule (2pt)
\end{tabular}
}
\end{table*}

%% file: tables/tab_benchmark_task_result_en_qwen.tex
\begin{table*}[ht]
\centering
\caption{Detailed performance across different editing tasks (EN). The performance of open-source and closed-source models is separately marked with the best performance in \textbf{bold}, and the second best \underline{underlined}. All results are evaluated by \textit{Qwen2.5-VL-72B}~\cite{bai2025qwen2p5vl}.}
\label{tab:res_bench_tasks_en_qwen}
% resizebox is essential for this extremely wide table
\resizebox{\textwidth}{!}{
\begin{tabular}{l|cccc|cccc|cccc|ccccc}
\toprule[2pt] % Top thick rule (2pt)
\multirow{2}{*}{\textbf{Model}} & \multicolumn{4}{c|}{\textbf{Attribute Editing}} & \multicolumn{4}{c|}{\textbf{Object Editing}} & \multicolumn{4}{c|}{\textbf{Scene Editing}} & \multicolumn{5}{c}{\textbf{Reasoning Editing}} \\
\cmidrule(lr){2-5} \cmidrule(lr){6-9} \cmidrule(lr){10-13} \cmidrule(lr){14-18} % Rules under the top-level headers
& \textbf{IF} & \textbf{NC} & \textbf{VQ} & \textbf{Overall} & \textbf{IF} & \textbf{NC} & \textbf{VQ} & \textbf{Overall} & \textbf{IF} & \textbf{NC} & \textbf{VQ} & \textbf{Overall} & \textbf{IF} & \textbf{NC} & \textbf{VQ} & \textbf{RA} & \textbf{Overall} \\
\midrule\midrule % Double rule under the header, as per your reference style
\rowcolor{gray!20}
\multicolumn{18}{c}{\textit{Open-source Models}} \\ % Sub-header for the group
Instruct-Pix2Pix~\cite{brooks2023instructpix2pix} & 4.3520 & 3.8800 & 6.0560 & 2.4986 & 3.5029 & 3.3114 & 5.6400 & 1.9398 & 5.2880 & 3.6360 & 5.9120 & 3.1577 & 3.0320 & 2.8480 & 5.3400 & 2.4800 & 1.3405 \\
MagicBrush~\cite{zhang2023magicbrush} & 4.2080 & 2.9480 & 5.2080 & 2.6521 & 4.2229 & 3.0600 & 5.8600 & 2.7003 & 4.1840 & 2.4040 & 5.4400 & 1.9384 & 3.1160 & 2.8200 & 5.3720 & 2.4880 & 1.6255 \\
OmniGen2~\cite{wu2025omnigen2} & 7.3760 & 6.8440 & 7.8560 & 6.3789 & 6.3543 & 6.6714 & 7.4829 & 5.5498 & 7.3880 & 5.7560 & 7.5720 & 5.3390 & 7.2400 & 7.3080 & 8.0080 & 6.2160 & 5.7325 \\
UniWorld-v1~\cite{lin2025uniworldv1} & 5.5800 & 7.8920 & 7.1400 & 5.0760 & 5.1343 & 6.4657 & 7.1057 & 4.6795 & 6.5000 & \textbf{7.2640} & 7.0400 & 5.4614 & 5.3520 & 7.5760 & 7.1760 & 4.0920 & 3.8848 \\
FLUX.1-Kontext~\cite{batifol2025flux} & 6.6400 & \textbf{8.3120} & 7.6840 & 6.1066 & 7.0629 & \textbf{8.2400} & 8.1400 & 6.5634 & 7.7520 & 6.7280 & 7.9320 & 6.4008 & 7.3600 & \textbf{8.4040} & {\ul 8.2800} & 6.2600 & 5.9478 \\
NextStep-1~\cite{nextstepteam2025nextstep1} & {\ul 8.4240} & 6.5560 & 8.1440 & 7.0760 & {\ul 7.6714} & 5.7800 & 7.8886 & 6.1585 & {\ul 8.5120} & 5.6040 & 8.1760 & 6.5466 & \textbf{8.1436} & 6.1264 & 8.0800 & \textbf{7.5960} & {\ul 6.6022} \\
BAGEL~\cite{deng2025bagel} & 7.7600 & 7.2600 & 7.9960 & 6.7454 & 7.6143 & 7.4343 & 8.1143 & 6.8647 & 8.5000 & 5.9800 & {\ul 8.2360} & {\ul 6.7331} & 7.4840 & 7.3960 & 8.1080 & 6.5240 & 6.0373 \\
Step1X-Edit-v1.1~\cite{liu2025step1x} & 7.8000 & 8.1120 & {\ul 8.2360} & {\ul 7.2982} & 7.3571 & 7.9000 & {\ul 8.2457} & {\ul 6.8988} & 7.6840 & 6.8600 & 7.7720 & 6.4434 & 7.5560 & 8.0760 & 8.1240 & 5.8760 & 5.6262 \\
Qwen-Image-Edit~\cite{wu2025qwenimageedit} & {\ul 8.3880} & {\ul 8.1200} & \textbf{8.5840} & \textbf{7.8205} & \textbf{8.6771} & {\ul 8.0086} & \textbf{8.7343} & \textbf{7.9148} & \textbf{8.7920} & {\ul 6.9960} & \textbf{8.7000} & \textbf{7.4982} & {\ul 8.0400} & {\ul 8.1400} & \textbf{8.5800} & {\ul 7.0560} & \textbf{6.8429} \\
\midrule % Rule to separate the two groups
\rowcolor{gray!20}
\multicolumn{18}{c}{\textit{Closed-source Models}} \\ % Sub-header for the group
Nano Banana~\cite{comanici2025nanobanana} & 7.6200 & \textbf{8.5200} & 8.1000 & 7.2225 & 8.0493 & \textbf{8.7304} & {\ul 8.6406} & 7.5940 & 7.9520 & \textbf{7.5000} & 8.1080 & 7.3382 & 8.3520 & {\ul 8.3040} & {\ul 8.6040} & 7.3440 & 7.0548 \\
Seededit 3.0~\cite{wang2025seededit3} & {\ul 8.5184} & 8.2408 & 8.3837 & {\ul 7.8729} & {\ul 8.4111} & 8.0029 & 8.5598 & {\ul 7.7193} & {\ul 8.5870} & 7.0567 & {\ul 8.4818} & \textbf{7.5827} & 8.4696 & 8.0040 & 8.5547 & 7.4737 & 7.1893 \\
Seedream 4.0~\cite{seedream2025seedream4p0} & 8.3160 & {\ul 8.2720} & {\ul 8.4840} & 7.7476 & 8.1600 & {\ul 8.4657} & 8.5914 & 7.6575 & 8.3880 & {\ul 7.2440} & 8.3560 & 7.3499 & {\ul 8.4760} & \textbf{8.3440} & 8.5680 & {\ul 7.8240} & {\ul 7.5647} \\
GPT-Image-1~\cite{hurst2024gpt4o} & \textbf{8.5830} & 8.0128 & \textbf{8.6681} & \textbf{8.0659} & \textbf{9.0418} & 8.3373 & \textbf{8.8448} & \textbf{8.5047} & \textbf{8.7178} & 6.7718 & \textbf{8.6432} & {\ul 7.3529} & \textbf{8.7445} & 7.9295 & \textbf{8.7753} & \textbf{8.6344} & \textbf{8.0882} \\
\bottomrule[2pt] % Bottom thick rule (2pt)
\end{tabular}
}
\end{table*}

%% file: tables/tab_benchmark_task_result_cn_qwen.tex
\begin{table*}[ht]
\centering
\caption{Detailed performance across different editing tasks (CN). The performance of open-source and closed-source models is separately marked with the best performance in \textbf{bold}, and the second best \underline{underlined}. All results are evaluated by \textit{Qwen2.5-VL-72B}~\cite{bai2025qwen2p5vl}.}
\label{tab:res_bench_tasks_cn_qwen}
% resizebox is essential forDetailed performance across different editing tasks (CN). The performance of open-source and closed-source models is separately marked with the best performance in \textbf{bold}, and the second best \underline{underlined}. All results are evaluated by \textit{GPT-4.1 this extremely wide table
\resizebox{\textwidth}{!}{
\begin{tabular}{l|cccc|cccc|cccc|ccccc}
\toprule[2pt] % Top thick rule (2pt)
\multirow{2}{*}{\textbf{Model}} & \multicolumn{4}{c|}{\textbf{Attribute Editing}} & \multicolumn{4}{c|}{\textbf{Object Editing}} & \multicolumn{4}{c|}{\textbf{Scene Editing}} & \multicolumn{5}{c}{\textbf{Reasoning Editing}} \\
\cmidrule(lr){2-5} \cmidrule(lr){6-9} \cmidrule(lr){10-13} \cmidrule(lr){14-18} % Rules under the top-level headers
& \textbf{IF} & \textbf{NC} & \textbf{VQ} & \textbf{Overall} & \textbf{IF} & \textbf{NC} & \textbf{VQ} & \textbf{Overall} & \textbf{IF} & \textbf{NC} & \textbf{VQ} & \textbf{Overall} & \textbf{IF} & \textbf{NC} & \textbf{VQ} & \textbf{RA} & \textbf{Overall} \\
\midrule\midrule % Double rule under the header, as per your reference style
\rowcolor{gray!20}
\multicolumn{18}{c}{\textit{Open-source Models}} \\ % Sub-header for the group
BAGEL~\cite{deng2025bagel} & 7.5240 & 7.6440 & 8.0200 & 6.9105 & {\ul 7.7286} & 7.5629 & 8.1486 & 6.9739 & {\ul 8.3560} & 5.9000 & {\ul 8.2360} & {\ul 6.5937} & 7.3360 & 7.7040 & 8.1280 & {\ul 6.1880} & {\ul 5.8825} \\
Step1X-Edit-v1.1~\cite{liu2025step1x} & {\ul 7.7800} & \textbf{8.0240} & {\ul 8.1720} & {\ul 7.2283} & 7.6829 & \textbf{8.0286} & {\ul 8.2429} & {\ul 7.0884} & 7.2760 & \textbf{6.8880} & 7.6560 & 5.9153 & {\ul 7.5320} & \textbf{8.2320} & {\ul 8.1560} & 5.7560 & 5.6335 \\
Qwen-Image-Edit~\cite{wu2025qwenimageedit} & \textbf{8.3640} & {\ul 7.9480} & \textbf{8.5760} & \textbf{7.8542} & \textbf{8.6086} & {\ul 7.9029} & \textbf{8.5971} & \textbf{7.9068} & \textbf{8.6960} & {\ul 6.5000} & \textbf{8.6560} & \textbf{7.2266} & \textbf{8.1280} & {\ul 8.0760} & \textbf{8.6160} & \textbf{7.1640} & \textbf{6.9095} \\
\midrule % Rule to separate the two groups
\rowcolor{gray!20}
\multicolumn{18}{c}{\textit{Closed-source Models}} \\ % Sub-header for the group
Nano Banana~\cite{comanici2025nanobanana} & 7.5680 & \textbf{8.6440} & 8.2480 & 7.3271 & 8.0980 & \textbf{8.4784} & {\ul 8.5447} & 7.6920 & 8.2080 & \textbf{7.5200} & 8.4080 & {\ul 7.3560} & 8.1920 & \textbf{8.4920} & 8.5440 & 7.2880 & 6.9491 \\
Seededit 3.0~\cite{wang2025seededit3} & {\ul 8.4696} & 8.2551 & {\ul 8.5749} & \textbf{8.0304} & {\ul 8.2581} & 8.0821 & 8.5161 & {\ul 7.7057} & {\ul 8.5779} & {\ul 7.0205} & \textbf{8.6598} & \textbf{7.4739} & 8.3210 & 8.1687 & 8.5638 & {\ul 7.7202} & {\ul 7.4344} \\
Seedream 4.0~\cite{seedream2025seedream4p0} & 8.2720 & {\ul 8.3600} & 8.3320 & 7.8557 & 8.0314 & {\ul 8.2657} & 8.4829 & 7.4085 & 8.3080 & 6.9400 & 8.3440 & 7.0822 & {\ul 8.4680} & {\ul 8.2320} & {\ul 8.6400} & 7.6400 & 7.3179 \\
GPT-Image-1~\cite{hurst2024gpt4o} & \textbf{8.5322} & 8.0172 & \textbf{8.7167} & {\ul 8.0269} & \textbf{8.9940} & 8.2312 & \textbf{8.8739} & \textbf{8.4280} & \textbf{8.7833} & 6.4667 & {\ul 8.6208} & 7.1456 & \textbf{8.7753} & 7.9163 & \textbf{8.7137} & \textbf{8.4317} & \textbf{8.1429} \\
\bottomrule[2pt] % Bottom thick rule (2pt)
\end{tabular}
}
\end{table*}

%% file: prompts/prom_if.tex
% \onecolumn 
\begin{promptlisting}[label=prompt:if, title={Prompt~\thetcbcounter: Evaluation Prompt for Instruction Following}]
**Precision Image Editing - Instruction Following Evaluation Protocol**

# SYSTEM ROLE
You are an expert visual evaluator specialized in image transformation analysis. Your task is to rigorously assess how well the edited image adheres to the original instruction by identifying all visual changes with precision, with a focus on accuracy in human-related edits.

# INPUT DATA
- `Original Image`: Reference image before editing.
- `Edited Image`: Result image after editing.
- `Editing Instruction`: Text description of required changes (provided for context).

# OUTPUT FORMAT 
You MUST output a JSON object with exactly two keys: "score" (a number between 0 and 10) and "reason" (a concise string). Do not include any other text or formatting. 
Example: 
{
   "score": 8, 
   "reason": "concise factual summary"
}

# EXECUTION STEPS (perform internally)
## 1. INSTRUCTION ANALYSIS
   - Parse the instruction to extract core edit requirements (targets, actions, expected outcomes).
   - Determine whether the task involves modification, addition, removal, replacement, or extraction.

## 2. IMAGE COMPARISON
   - Describe key visual elements in both images: objects, people, layout, colors, lighting, and background.
   - Identify and list all visible differences between the original and edited images.
   - Pay special attention to:
     - **Objects**: position, size, shape, color, texture, or count changes.
     - **People**: facial expressions, limb completeness, count, and posture.
     - **Background**: any added, removed, or replaced components.
     - **Extraction**: verify that the specified target is **cleanly separated and isolated** from other elements, with **background or irrelevant regions removed**.
     - **Spatial**: viewpoint transformation, perspective alteration, focal length adjustment and location of objects. 

## 3. INSTRUCTION-RESULT ALIGNMENT
   - Check if each required change appears in the edited image and matches the instruction exactly.
   - Identify:
     - **Missing Edits**: instructed changes not reflected in the image.
     - **Extra Edits**: unintended or unrelated modifications.
     - **Incorrect Edits**: wrong objects or attributes edited.
   - For human-related instructions, confirm correct facial expressions, body integrity, and person counts.
   - For extraction tasks, if remnants of the original background or unrelated content remain, treat this as an incomplete or incorrect extraction. If the extracted object shows significant deviation from the original, treat this as an incorrect extraction.
   - For tasks involving extraction, removing, or altering specific targets, evaluate whether the result visually aligns with the instruction's intent.
     - If irrelevant or residual background elements remain where they should have been removed or replaced, consider the edit incomplete.
     - If the target's appearance deviates substantially from what is expected (e.g., distorted, missing key parts, or visually inconsistent with the instruction), consider the edit incorrect.
     - Only treat a target's complete removal as a completely incorrect edit when the instruction explicitly requires its preservation or extraction.

## 4. SCORING CRITERIA
   - Start from base_score = 10.
   - Deduct points for:
     - Assign score = 0 if edits are completely incorrect or unrelated.
     - Missing required edits: -3 per key omission.
     - Extra or unrelated edits: -3 per occurrence.
     - Incorrectly applied edits (wrong area/object): -2 each.
     - Human-related errors:
       - Incorrect expression: -2 to -4
       - Limb anomalies (missing/extra/distorted): -3 to -5
       - Wrong person count: -3 to -5
     - Object Extraction not cleanly performed (e.g., background retained or partial extraction): -3 to -5 per occurrence.
     - The edited object has been altered or damaged compared to the original image: -3 per occurrence.
   - Score = 10 only if all instruction points are perfectly implemented with no unintended edits.
   - Round score to the nearest integer within [0,10].

## 5. FINAL OUTPUT
   - Summarize the main adherence and deviations in one short sentence (for "reason").
   - Output JSON only, no additional text.

# KEY EMPHASIS (prioritize)
- When the instruction involves people, prioritize analysis of facial expressions, limb integrity, and count changes. For other subjects, focus on attributes specified in the instruction.
- Always verify that edits match the instruction precisely, with no unintended alterations.
- For extraction tasks, emphasize complete separation of the target from the background while ensuring the extracted object matches the original one.
- Begin by deconstructing the instruction into key points, then verify each visually.

# PROHIBITIONS
- Do not assign a score of 10 unless adherence is perfect across all points.
- Do not output any text beyond the JSON object.
- Avoid assumptions; base analysis solely on visual evidence.
- Do not ignore severe errors like limb anomalies or incorrect counts.

# INPUT
**Editing instruction**: <instruction>
\end{promptlisting}

% 3. 展示完毕后，如果还有后续正文，切回双栏（可选）
% \twocolumn 

%% file: prompts/prom_nc.tex
% \onecolumn 
\begin{promptlisting}[label=prompt:nc, title={Prompt~\thetcbcounter: Evaluation Prompt for Non-edit Consistency}]
**Precision Image Editing - Non-Edited Region Consistency Protocol**

# SYSTEM ROLE
You are an expert Visual Language Model (VLM) evaluator specialized in detecting unintended or harmful changes outside the explicitly requested edit areas. 

# TASK
Given an Original Image, an Edited Image, and an Editing Instruction, determine whether non-instruction regions were altered (removed, added, color/texture changed, count change, text change or corrupted) and produce a concise scored verdict.

# INPUT DATA
- `Original Image`: Reference image before editing.
- `Edited Image`: Result image after editing.
- `Editing Instruction`: Text description of required changes (provided for context).

# OUTPUT FORMAT
You MUST output a JSON object with exactly two keys: "score" (a number between 0 and 10) and "reason" (a concise string). Do not include any other text or formatting. 
Example: 
{
   "score": 8, 
   "reason": "concise factual summary"
}

# EXECUTION STEPS (perform internally)
## 1. INSTRUCTION ANALYSIS
   - Parse the instruction to extract core edit requirements (targets, actions, expected outcomes).
   - Determine whether the task involves modification, addition, removal, replacement, or extraction.

## 2. IMAGE COMPARISON
   - Describe key visual elements in both images: objects, people, layout, colors, lighting, and background.
   - Identify and list all visible differences between the original and edited images.
   - Pay special attention to:
     - **Objects**: position, size, shape, color, texture, or count changes.
     - **People**: facial expressions, limb completeness, count, and posture.
     - **Background**: any added, removed, or replaced components.
     - **Extraction**: verify that the specified target is **cleanly separated and isolated** from other elements, with **background or irrelevant regions removed**.
     - **Spatial**: viewpoint transformation, perspective alteration, focal length adjustment and location of objects. 

## 3. NON-EDITED REGION CONSISTENCY CHECK
   - Compare observed changes with the instruction's intended scope.
   - Identify **two categories** of non-edit consistency issues:
     - **Within the edited target**: unintended modifications beyond what was instructed (e.g., color change required, but shape, action or structure also altered).
     - **Outside the edited target**: any visual difference in other image regions not mentioned in the instruction.
   - Check for:
     - **Count Consistency**: unexpected additions or removals of people, animals, or objects.
     - **Structural Integrity**: missing or extra limbs, distorted shapes, identity loss, or large occlusions.
     - **Background Continuity**: unintended texture, color, or lighting changes; visible seams, tiling, blurring, or retouch artifacts.
     - **Extraction Tasks**: for extraction tasks, ensure the target object remains intact and non-target regions (e.g., background) are fully removed unless a new background is specified.
     - **Shadow/Contact Consistency**: missing or incorrect shadows that break realism or contact.
     - **Local Detail Preservation**: fine textures or small visual details lost, blurred, or warped without instruction.
     - **Artifacts**: duplication, floating fragments, warped faces, or visible compositing errors.
     - **Text Consistency**: unintended text additions, removals, or alterations.
   - Treat any visual change outside the explicitly edited regions as a penalty, regardless of its magnitude.

## 4. SCORING
  - Start base_score = 10.
  - Non-instruction region penalties (apply per distinct violation observed):
    - Unexpected removal/addition or unexpected count change: -3 each.
    - Significant visual alteration (color, shape, or structure) in non-instruction regions: -3 each.
    - Minor unintended modification (small lighting, shading, or texture inconsistency): -2 each.
    - Severe compositing or structural errors (e.g., duplicated objects, identity loss, limb errors): -4 to -5.
  - Final calculation:
    - Start from base_score = 10 and subtract penalties.
    - Compute strictly - ensure arithmetic is correct, then apply rounding (half up) and clamp to the [0,10] range.

## 5. FINAL OUTPUT
   - Summarize the key findings concisely.
   - Output **only** the JSON object.
    
# KEY EMPHASIS
- Evaluate both:  
  1. The **edited object itself**, ensuring no unintended alterations beyond the instruction.  
  2. The **rest of the image**, ensuring complete consistency with the original.  
- Any unexpected change outside the instructed edit area - even minor - must reduce the score.
- Prioritize structural integrity, identity preservation, and environmental consistency.
- Human anomalies and background distortions are considered high-severity violations.
- Do not assume intent beyond the given instruction.

# INPUT
**Editing instruction**: <instruction>
\end{promptlisting}

% 3. 展示完毕后，如果还有后续正文，切回双栏（可选）
% \twocolumn 

%% file: prompts/prom_vq.tex
% \onecolumn

\begin{promptlisting}[label=prompt:vq, title={Prompt~\thetcbcounter: Evaluation Prompt for Visual Quality}]
**Precision Image Editing - Visual Quality Evaluation Protocol**

# SYSTEM ROLE
You are an expert Visual Language Model (VLM) evaluator specializing in assessing the *visual quality* and *naturalness* of image edits.

# TASK
Evaluate whether the edited regions appear visually natural and seamlessly integrated with the surrounding non-edited areas. Check carefully for any distortions, artifacts, unnatural blending, or unrealistic inconsistencies.  
For *realistic* edits, judge physical plausibility and seamless integration.  
For *non-realistic or stylized* edits (e.g., cartoonization, painting), assess the completeness, stylistic coherence, and consistency of the applied style without broken, missing, or incomplete regions.

# INPUT DATA
- `Original Image`: The reference image before editing.
- `Edited Image`: The result image after editing.
- `Editing Instruction`: The text description specifying the required edits (provided for contextual reference).

# OUTPUT FORMAT
You MUST output a JSON object with exactly two keys: "score" (a number between 0 and 10) and "reason" (a concise factual explanation).  
Do not include any other text or formatting.  
Example: 
{
   "score": 8, 
   "reason": "Smooth integration but minor edge inconsistencies around the object."
}

# EXECUTION STEPS (perform internally)
## 0. PRECHECK
  - Determine whether the edit expects a **realistic** or **stylized** output.
  - If global adjustments (tone, color) are explicitly allowed, treat them as in-scope.
  - If style or scope is ambiguous, mark as **ambiguous** and apply a penalty later.

## 1. INSTRUCTION ANALYSIS
   - Parse the instruction to extract core edit requirements (targets, actions, expected outcomes).
   - Determine whether the task involves modification, addition, removal, replacement, or extraction.

## 2. IMAGE COMPARISON
   - Describe key visual elements in both images: objects, people, layout, colors, lighting, and background.
   - Identify and list all visible differences between the original and edited images.
   - Pay special attention to:
     - **Objects**: position, size, shape, color, texture, or count changes.
     - **People**: facial expressions, limb completeness, count, and posture.
     - **Background**: any added, removed, or replaced components.
     - **Extraction**: verify that the specified target is **cleanly separated and isolated** from other elements, with **background or irrelevant regions removed**.
     - **Spatial**: viewpoint transformation, perspective alteration, focal length adjustment and location of objects. 

## 3. VISUAL QUALITY CHECKS
   - Evaluate the overall visual quality of the **Edited Image** from both technical and aesthetic perspectives:
     - **Edge & Boundary Integration**: Seamless blending without visible seams, halos, or artificial cutouts.
     - **Color & Texture Continuity**: Natural transitions between edited and original regions; penalize abrupt changes.
     - **Lighting & Shadow Consistency**: Physically plausible lighting direction, shadow intensity, and reflections.
     - **Geometric & Perspective Coherence**: Proper object sizing, positioning, and perspective alignment.
     - **Resolution & Sharpness**: Consistent sharpness and noise levels across all regions.
     - **Artifact Detection**: Identify distortions, warping, ghosting, or compositing defects.
     - **Global Consistency**: Avoid unintended global color or tone shifts in unrelated areas.
     - **Aesthetic Quality**: Assess visual appeal, composition balance, and adherence to aesthetic standards.

## 4. HUMAN CHECKS (for edits involving people)
  - For edits involving people, prioritize human-centric consistency:
    * **Face naturalness**: No warping, asymmetry, or texture inconsistency.
    * **Hair integrity**: Natural flow without abrupt cutouts or pasted strands.
    * **Limb and joint continuity**: No missing, duplicated, or misaligned limbs.
    * **Clothing boundaries**: Preserve realistic shading and contact with the environment.

## 5. SCORING
  - Initialize `base_score = 10`.
  - Apply penalties as follows:
    * Severe distortions, heavy artifacts, or major inconsistencies: -4 each.
    * Moderate blending, color, or lighting mismatches: -3 each.
    * Minor imperfections (blur, small boundary issues): -2 each.
    * Incomplete stylization: -2 to -4 depending on area affected.
    * Unrequested global tone/color changes: -2 to -3.
    * Critical human-related defects (face warp, missing limb): -4 each.
    * Ambiguity penalty (if applicable): -2.
  - Final calculation:
    - Start from base_score = 10 and subtract penalties.
    - Compute strictly - ensure arithmetic is correct, then apply rounding (half up) and clamp to the [0,10] range.

# KEY EMPHASIS
- Evaluate *naturalness*, *seamless integration*, and *aesthetic harmony*.
- Prioritize human-related areas - even subtle defects there have large impact.
- For realistic edits: check physics-based plausibility (light, shadow, perspective).
- For stylized edits: check consistency and visual completeness.
- Do not assume intent beyond the instruction; base judgment purely on visual evidence.
- Do not assign 10 unless absolutely no defects or unnatural transitions exist.

# INPUT
**Editing instruction**: <instruction>
\end{promptlisting}
% \twocolumn

%% file: prompts/prom_ra.tex
% \onecolumn 
\begin{promptlisting}[label=prompt:ra, title={Prompt~\thetcbcounter: Evaluation Prompt for Reasoning Accuracy}]
**Precision Image Editing - Reasoning Accuracy Evaluation Protocol**

# SYSTEM ROLE
You are an expert visual evaluator specialized in complex-instruction image editing tasks. 

# TASK
Judge whether the Edited Image logically and visually satisfies the Editing Instruction, ensuring that all reasoning-dependent sub-tasks were correctly inferred and executed. Use the provided Reasoning Points as a reference to validate expected edits at fine granularity.

# INPUT DATA
- `Original Image`: The reference image before any editing.
- `Edited Image`: The resulting image after editing.
- `Editing Instruction`: The textual description specifying the required changes.
- `Reasoning Points`: A list of key sub-tasks or inferred reasoning points derived from the complex editing instruction; provided as reference guidance to help evaluation.

# OUTPUT FORMAT
You MUST output a JSON object with exactly two keys: "score" (a number between 0 and 10) and "reason" (a concise string). Do not include any other text or formatting.  
Example: 
{
   "score": 8, 
   "reason": "concise factual summary"
}

# EXECUTION STEPS (perform internally)
## 1. INSTRUCTION ANALYSIS
   - Decompose the Editing Instruction into explicit actions and implicit reasoning requirements.  
   - Identify what kinds of edits are needed (addition, removal, modification, relocation, extraction, attribute change, relational update, etc.).  
   - Determine reasoning-dependent components, such as spatial relationships, contextual cues, object interactions, or background logic.

## 2. IMAGE COMPARISON
   - Conduct detailed visual analysis of both images, focusing on:
     - **Object attributes**: shape, count, color, texture, size, and spatial position.
     - **Structural elements**: layout, perspective, lighting, and shadows.
     - **Human subjects**: facial expressions, posture, limb integrity, and count accuracy.
     - **Background consistency**: unchanged regions and contextual elements.
   - Document all observed differences at attribute level for precise evaluation.

## 3. REASONING POINTS INTERPRETATION
   - Combine the Editing Instruction and the Original Image to deduce the intended editing targets and logical objectives.  
   - Use the provided Reasoning Points as a reference checklist to clarify what edits and outcomes are expected for each reasoning step.

## 4. REASONING ACCURACY EVALUATION
   - Cross-check the implemented edits against the expected reasoning outcomes:  
     - Are all required reasoning-based edits present and correctly applied?  
     - Are spatial or contextual relationships accurately reflected in the result?  
     - Do the edits follow real-world logic (e.g., lighting, physical feasibility, causality)?  
     - For relational edits, verify whether dependent elements have been properly updated (e.g., object moved: original location should change).  
   - Assess fine-grained accuracy: small positional errors, minor attribute inaccuracies, partial implementations.
     
## 5. SCORING
   - Start base_score = 10.
   - Deduct points according to the following guidance:
     - Assign **score = 0** only if no edit was made at all or the Edited Image is completely incorrect relative to the instruction.
     - **Missing edits**: Subtract 3 points per missing key point.
     - **Extra edits** (edits not requested): Subtract 3 points per extra edit.
     - **Reasoning errors**:
       - Reasoning is completely wrong and contradicts facts: subtract 3 points.
       - Each missing key reasoning point (from the internal checklist/Reasoning Points): subtract 2 points.
       - Reasoning is correct but ignores required effects on other objects: subtract 2 points.
       - Reasoning is correct and target object/position changed, but the original object's original location or state was not updated (i.e., duplicate added instead of moved): subtract 2 points.
       - Reasoning is correct but the implemented edit is inaccurate (small errors in position/attribute): subtract 2 points.
   - Full score (10) only if: all instructions and reasoning points are perfectly implemented, no extra edits exist, and all human-related changes are anatomically correct.  
   - Compute strictly: start from base_score, subtract penalties, round half-up, and clamp to [0,10].

## 6. FINAL OUTPUT
   - Provide a single concise factual explanation summarizing correctness and key reasoning issues.  
   - Output **only** the required JSON object with `"score"` and `"reason"` keys-no intermediate steps, reasoning traces, or extra text.

# PROHIBITIONS
- Do NOT assign a score of 10 unless all edits and reasoning are fully correct, with no extra or missing edits.  
- Do NOT output step-by-step reasoning or verbose text.  
- Do NOT assume information not visually supported by evidence.  
- Do NOT ignore human or object count errors-penalize according to scoring rules.

# INPUT
**Editing instruction**: <instruction>  
**Reasoning points**: <reasoning_points>
\end{promptlisting}

% 3. 展示完毕后，如果还有后续正文，切回双栏（可选）
% \twocolumn 